\def\year{2020}\relax
\documentclass[letterpaper]{article} % DO NOT CHANGE THIS
\usepackage{aaai20}  % DO NOT CHANGE THIS
\usepackage{times}  % DO NOT CHANGE THIS
\usepackage{helvet} % DO NOT CHANGE THIS
\usepackage{courier}  % DO NOT CHANGE THIS
\usepackage[hyphens]{url}  % DO NOT CHANGE THIS
\usepackage{graphicx} % DO NOT CHANGE THIS
\usepackage{graphicx}  % DO NOT CHANGE THIS
\usepackage{amsfonts}
\usepackage{amsmath}
\usepackage{amsthm}
\usepackage{algorithm}
\usepackage[noend]{algpseudocode}
\usepackage[title]{appendix}
\usepackage{graphicx}
\usepackage{subcaption}
\usepackage{booktabs}

\newtheorem{theorem}[]{Theorem}
\newtheorem{lemma}[]{Lemma}

\newtheorem{corollary}[]{Corollary}

\newcommand{\newciteauthor}[1]{\citeauthor{#1} (\citeyear{#1})}

\title{Shapley Q-value: A Local Reward Approach to Solve Global Reward Games}
\author{
  Jianhong Wang$^{12}\thanks{The first two authors have equal contributions. Correspondence to \texttt{jianhong.wang16@imperial.ac.uk}.}$, Yuan Zhang$^{3*}$, Tae-Kyun Kim$^2$, Yunjie Gu$^{1}$\\
  $^1$ Control and Power Research Group, Imperial College London, UK\\
  $^2$ Imperial Computer Vision and Learning Lab, Imperial College London, UK\\
  $^3$ Laiye Network Technology Co.Ltd., China
}

\begin{document}

\maketitle

\begin{abstract}
Cooperative game is a critical research area in the multi-agent reinforcement learning (MARL). Global reward game is a subclass of cooperative games, where all agents aim to maximize the global reward. Credit assignment is an important problem studied in the global reward game. Most of previous works stood by the view of non-cooperative-game theoretical framework with the shared reward approach, i.e., each agent being assigned a shared global reward directly. This, however, may give each agent an inaccurate reward on its contribution to the group, which could cause inefficient learning. To deal with this problem, we i) introduce a cooperative-game theoretical framework called extended convex game (ECG) that is a superset of global reward game, and ii) propose a local reward approach called Shapley Q-value. Shapley Q-value is able to distribute the global reward, reflecting each agent's own contribution in contrast to the shared reward approach. Moreover, we derive an MARL algorithm called Shapley Q-value deep deterministic policy gradient (SQDDPG), using Shapley Q-value as the critic for each agent. We evaluate SQDDPG on Cooperative Navigation, Prey-and-Predator and Traffic Junction, compared with the state-of-the-art algorithms, e.g., MADDPG, COMA, Independent DDPG and Independent A2C. In the experiments, SQDDPG shows a significant improvement on the convergence rate. Finally, we plot Shapley Q-value and validate the property of fair credit assignment.
\end{abstract}

\section{Introduction}
\label{sec:introduction}
    Cooperative game is a critical research area in multi-agent reinforcement learning (MARL). Many real-life tasks can be modeled as cooperative games, e.g., the coordination of autonomous vehicles \cite{keviczky2007decentralized}, autonomous distributed logistics \cite{schuldt2012multiagent} and search-and-rescue robots \cite{koes2006constraint,ramchurn2010decentralized}. Global reward game \cite{chang2004all} is a subclass of cooperative games where agents aim to maximize the global reward. In this game, credit assignment is an important problem, which targets on finding a method to distribute the global reward. There are two categories of approaches to solve out this problem, namely shared reward approach (also known as shared reward game or fully cooperative game) \cite{NIPS2016_6398,omidshafiei2018learning,kim2018learning} and local reward approach \cite{panait2005cooperative}. The shared reward approach directly assigns the global reward to all agents. The local reward approach, on the other hand, distributes the global reward according to each agent's contribution, and turns out to have superior performances in many tasks \cite{foerster2018counterfactual,nguyen2018credit}.  
    
    Whatever approach is adopted, a remaining open question is whether there is an underlying theory to explain credit assignment. Conventionally, a global reward game is built upon non-cooperative game theory, which primarily aims to find Nash equilibrium as the stable solution \cite{osborne1994course,basar1999dynamic}. This formulation can be extended to a dynamic environment with infinite horizons via stochastic game \cite{shapley1953stochastic}. However, Nash equilibrium focuses on the individual reward and has no explicit incentives for cooperation. As a result, the shared reward function has to be applied to force cooperation, which could be used as a possible explanation to the shared reward approach, but not the local reward approach.
    
    In our work, we introduce and investigate the cooperative game theory (or the coalitional game theory) \cite{chalkiadakis2011computational} in which local reward approach becomes rationalized. In cooperative game theory, the objective is dividing coalitions and binding agreements among agents who belong to the same coalition. We focus on convex game (CG) which is a typical game in cooperative game theory featuring the existence of a stable coalition structure with an efficient payoff distribution scheme (i.e., a local reward approach) called core. This payoff distribution is equivalent to the credit assignment, thereby the core rationalizes and well explains the local reward approach \cite{peleg2007introduction}.
    
    Referring to the previous work \cite{suijs1999stochastic}, we extend CG to an infinite-horizon scenario, namely extended CG (ECG). In addition, we show that a global reward game is equivalent to an ECG with the grand coalition and an efficient payoff distribution scheme. Furthermore, we propose Shapley Q-value, extending Shapley value (i.e., an efficient payoff distribution scheme) for the credit assignment in an ECG with the grand coalition. Therefore, it results in a conclusion that Shapley Q-value is able to work in a global reward game. Finally, we derive an algorithm called Shapley Q-value deep deterministic policy gradient (SQDDPG) according to the actor-critic framework \cite{konda2000actor} to learn decentralized policies with centralized critics (i.e., Shapley Q-values). SQDDPG is evaluated on the environments such as Cooperative Navigation, Prey-and-Predator \cite{lowe2017multi}, and Traffic Junction \cite{NIPS2016_6398}, compared with the state-of-the-art baselines, e.g., MADDPG \cite{lowe2017multi}, COMA \cite{foerster2018counterfactual}, Independent DDPG \cite{lillicrap2015continuous} and Independent A2C \cite{sutton2018reinforcement}.
    
%===============================================================================

\section{Related Work}
\label{sec:related_work}
    \subsubsection{Multi-agent Learning} Multi-agent learning refers to a category of methods that tackle the games with multiple agents such as cooperative games. Among these methods, we only focus on using reinforcement learning to deal with a cooperative game, which is called multi-agent reinforcement learning (MARL). Incredible progresses have recently been made on MARL. Some researchers \cite{NIPS2016_6398,kim2018learning,das2018tarmac} focus on distributed executions, which allow communications among agents. Others \cite{chang2004all,foerster2018counterfactual,nguyen2018credit,lowe2017multi,iqbal2018actor} consider decentralized executions, where no communication is permitted during the execution. Nevertheless, all of them study on centralized critics, which means information can be shared on the value function during training. In our work, we pay our attention to the decentralized execution and the centralized critic.
    
    \subsubsection{Cooperative Game} As opposed to competing with others, agents in a cooperative game aim to cooperate to solve a joint task or maximize the global payoff (also known as the global reward) \cite{chalkiadakis2011computational}. \newciteauthor{shapley1953stochastic} proposed a non-cooperative game theoretical framework called stochastic game, which models the dynamics of multiple agents in zero-sum game with infinite horizons. \newciteauthor{hu1998multiagent} introduced a general-sum stochastic game theoretical framework, which generalises the zero-sum game. To force cooperation under this framework, potential function \cite{monderer1996potential} was applied such that each agent shares the same objective, namely global reward game \cite{chang2004all}. In this paper, we use cooperative game theory whereas the existing cooperative game framework are built under the non-cooperative game theory. Our framework gives a new view on the global reward game and well explains the reason why credit assignment is important. We show that the global reward game is a subclass of our framework if we interpret that the agents in a global reward game forms a grand coalition (i.e., the group including the whole agents). Under our framework, it is more rational to use a local reward approach to distribute the global reward.
    
    \subsubsection{Credit Assignment} Credit assignment is a significant problem that has been studied in cooperative games for a long period. There are two sorts of credit assignment approaches, i.e., shared reward approach and local reward approach. The shared reward approach directly assigns each agent the global reward \cite{NIPS2016_6398,kim2018learning,das2018tarmac,lowe2017multi}. We show that this is actually equivalent to distributing the global reward equally to individual agents. The global reward game with this credit assignment scheme is also called shared reward game (also known as fully cooperative game) \cite{panait2005cooperative}. However, \newciteauthor{wolpert2002optimal} claimed 
    that the shared reward approach does not give each agent the accurate contribution. Thus, it may not perform well in difficult problems. This motivates the study on the local reward approach, which can distribute the global reward to agents according to their contributions. The existing question is how to quantify the contributions. To investigate the answer to this question, \newciteauthor{chang2004all} attempted using Kalman filter to infer the contribution of each agent. Recently, \newciteauthor{foerster2018counterfactual} and \newciteauthor{nguyen2018credit} modelled the marginal contributions inspired by the reward difference \cite{wolpert2002optimal}. Under our proposed framework (i.e., ECG), we propose a new method called Shapley Q-value to learn a local reward. This method is extended from the Shapley value \cite{shapley1953value}. It is theoretically guaranteed to distribute the global reward fairly. Although Shapley value can be regarded as the expectation of the marginal contributions, it is different from the previous works \cite{foerster2018counterfactual,nguyen2018credit}: it considers all possible orders of agents to form a grand coalition, which has not been mentioned in these works.
    
%===============================================================================

\section{Preliminaries}
\label{sec:background}
    
    \subsection{Convex Game}
    \label{subsec:convex game}
        Convex game (CG) is a typical transferable utility game in the cooperative game theory. The definitions below are referred to the textbook \cite{chalkiadakis2011computational}. A CG is formally represented as $\Gamma=\langle \mathcal{N}, v \rangle$, where $\mathcal{N}$ is the set of agents and $v$ is the value function to measure the profits earned by a coalition (i.e., a group). $\mathcal{N}$ itself is called the grand coalition. The value function $v: 2^{\mathcal{N}} \rightarrow \mathbb{R}$ is a mapping from a coalition $\mathcal{C} \subseteq \mathcal{N}$ to a real number $v(\mathcal{C})$. In a CG, its value function satisfies two properties, i.e., 1) $v(\mathcal{C} \cup \mathcal{D}) \geq v(\mathcal{C}) + v(\mathcal{D}), \forall \mathcal{C}, \mathcal{D} \subset \mathcal{N}, \mathcal{C} \cap \mathcal{D} = \phi$; 2) the coalitions are independent. The solution of a CG is a tuple $(\mathcal{CS}, \textbf{x})$, where $\mathcal{CS} = \{ \mathcal{C}_{1}, \mathcal{C}_{2}, ..., \mathcal{C}_{m} \}$ is a coalition structure and $\textbf{x}=(\text{x}_{i})_{i \in \mathcal{N}}$ indicates the payoff (i.e., the local reward) distributed to each agent, which satisfies two conditions, i.e., 1) $\text{x}_{i} \geq 0, \forall i \in \mathcal{N}$; 2) $\textbf{x}(\mathcal{C}) \leq v(\mathcal{C}), \forall \mathcal{C} \subseteq \mathcal{CS}$, where $\textbf{x}(\mathcal{C}) = \sum_{i \in \mathcal{C}} \text{x}_{i}$. $\mathcal{CS}_{\mathcal{N}}$ denotes the set of all possible coalition structures. The \textit{core} is a stable solution set of a CG, which can be defined mathematically as $\texttt{core}(\Gamma) = \{ (\mathcal{C}, \textbf{x}) | \textbf{x}(\mathcal{C}) \geq v(\mathcal{C}), \forall \mathcal{C} \subseteq \mathcal{N} \}$. The core of a CG ensures a reasonable payoff distribution and inspires our work on credit assignment in MARL.
    
    \subsection{Shapley Value}
    \label{subsec:shapley value}
        Shapley value \cite{shapley1953value} is one of the most popular methods to solve the payoff distribution problem for a grand coalition \cite{fatima2008linear,michalak2013efficient,faigle1992shapley}. Given a cooperative game $\Gamma=(\mathcal{N}, v)$, for any $\mathcal{C} \subseteq \mathcal{N} \backslash \{i\}$ let $\delta_{i}(\mathcal{C}) = v(\mathcal{C} \cup \{i\}) - v(\mathcal{C})$ be a marginal contribution, then the Shapley value of each agent $i$ can be written as: 
        \begin{equation}
            \text{Sh}_{i}(\Gamma)= \sum_{\mathcal{C} \subseteq \mathcal{N} \backslash \{i\}} \frac{|\mathcal{C}|!(|\mathcal{N}|-|\mathcal{C}|-1)!}{|\mathcal{N}|!} \cdot \delta_{i}(\mathcal{C}).
            \label{eq: shapley_value_org}
        \end{equation}
        Literally, Shapley value takes the average of marginal contributions of all possible coalitions, so that it satisfies: 1) efficiency: $\textbf{x}(\mathcal{N})=v(\mathcal{N})$; 2) dummy player: if an agent $i$ has no contribution, then $x_{i}=0$; and 3) symmetry: if $i$-th and $j$-th agents have the same contribution, then $x_{i} = x_{j}$ \cite{chalkiadakis2011computational}. All these properties form the fairness. As we can see from Eq.\ref{eq: shapley_value_org}, if we calculate Shapley value for an agent, we have to consider $2^{|\mathcal{N}|}-1$ possible coalitions that the agent could join in to form a grand coalition, which causes the computational catastrophe. To mitigate this issue, we propose an approximation in the scenarios with infinite horizons called approximate Shapley Q-value which is introduced in the next section.
    
    \subsection{Multi-agent Actor-Critic}
    \label{subsec:multi-agent actor-critic}
        Different from the value based method, i.e., Q-learning \cite{watkins1992q}, policy gradient \cite{williams1992simple} directly learns the policy by maximizing $J(\theta) = \mathbb{E}_{s \sim \rho^{\pi}, a \sim \pi_{\theta}}[r(s, a)]$, where $r(s, a)$ is the reward of an arbitrary state-action pair. Since the gradient of $J(\theta)$ w.r.t. $\theta$ cannot be directly calculated, policy gradient theorem \cite{sutton2018reinforcement} is used to approximate the gradient such that $\nabla_{\theta} J(\theta) = \mathbb{E}_{s \sim \rho^{\pi}, a \sim \pi_{\theta}}[Q^{\pi}(s, a) \nabla_{\theta} \log \pi_{\theta}(a|s)]$. In the actor-critic framework \cite{konda2000actor} (that is derived from the policy gradient theorem), $\pi_{\theta}(a|s)$ is called actor and $Q^{\pi}(s, a)$ is called critic. Additionally, $Q^{\pi}(s, a) = \mathbb{E}_{\pi}[\sum_{t=1}^{\infty} \gamma^{t-1} r(s_{t}, a_{t})|s_{1}\text{ = }s, a_{1}\text{ = }a]$. Extending to the multi-agent scenarios, the gradient of each agent $i$ can be represented as $\nabla_{\theta_{i}} J(\theta_{i}) = \mathbb{E}_{s \sim \rho^{\pi}, \textbf{a} \sim \pi_{\theta}}[Q_{i}^{\pi}(s, a_{i}) \nabla_{\theta_{i}} \log \pi_{\theta_{i}}^{i}(a_{i}|s)]$. $Q_{i}^{\pi}(s, a_{i})$ can be regarded as the estimation of the contribution of each agent $i$. If the deterministic policy \cite{silver2014deterministic} needs to be learned in MARL problems, we can reformulate the approximate gradient of each agent as $\nabla_{\theta_{i}}J(\theta_{i}) = \mathbb{E}_{s \sim \rho^{\mu}}[\nabla_{\theta_{i}} \mu_{\theta_{i}}^{i}(s) \nabla_{a_{i}} Q_{i}^{\mu}(s, a_{i})|_{a_{i}=\mu_{\theta_{i}}^{i}(s)}]$. In this work, we applies this formulation to learn the deterministic policy for each agent.
    
%===============================================================================

\section{Our Work}
\label{sec:our_work}
    In this section, we (i) extend convex game (CG) with the infinite horizons and decisions, namely extended convex game (ECG) and show that a global reward game is equivalent to an ECG with the grand coalition and an efficient distribution scheme, (ii) show that the shared reward approach is an efficient distribution scheme in an ECG with the grand coalition, (iii) propose Shapley Q-value by extending and approximating Shapley value to distribute the credits in a global reward game, because it can accelerate the convergence rate compared with shared reward approach, and (iv) derive an MARL algorithm called Shapley Q-value deep deterministic policy gradient (SQDDPG), using the Shapley Q-value as each agent's critic.
    
    \subsection{Extended Convex Game}
    \label{subsec:esg}
        Referring to the previous work \cite{suijs1999stochastic,chalkiadakis2004bayesian}, we extend the CG to the scenarios with infinite horizons and decisions, named as extended CG (ECG). The set of joint actions of agents is defined as $\mathcal{A} = \times_{i \in \mathcal{N}} \mathcal{A}_{i}$, where $\mathcal{A}_{i}$ is the feasible action set for each agent $i$. $\mathcal{S}$ is the set of possible states in the environment. The dynamics of the environment are defined as $Pr(s'|s, \textbf{a})$, where $s, s' \in \mathcal{S}$ and $\textbf{a} \in \mathcal{A}$. Inspired by \newciteauthor{nash1953two}, we construct the ECG by two stages. In the stage 1, an oracle arranges the coalition structure and contracts the cooperation agreements, i.e., the credit assigned to an agent for his optimal long-term contribution if he joins in some coalition. We assume that this oracle can observe the whole environment and be familiar with each agent's feature. In the stage 2, after joining in the allocated coalition, each agent will further make a decision by $\pi_{i}(a_{i}|s)$ to maximize the social value of its coalition, so that the optimal social value of each coalition and individual credit assignment can be achieved, where $a_{i} \in \mathcal{A}_{i}$ and $s \in \mathcal{S}$. Mathematically, the optimal value of a coalition $\mathcal{C} \in \mathcal{CS}$ can be written as $\max_{\pi_{\mathcal{C}}} v^{\pi_{\mathcal{C}}}(\mathcal{C}) = \mathbb{E}_{\pi_{\mathcal{C}}}[\sum_{t=1}^{\infty} \gamma^{t-1} r_{t}(\mathcal{C})]$; $\pi_{\mathcal{C}} = \times_{i \in \mathcal{C}} \pi_{i}$; $r_{t}(\mathcal{C})$ is the reward gained by coalition $\mathcal{C}$ at each time step. According to the property (1) of the CG aforementioned, the formula $\max_{\pi_{\mathcal{C} \cup \mathcal{D}}} v^{\pi_{\mathcal{C} \cup \mathcal{D}}}(\mathcal{C} \cup \mathcal{D}) \geq \max_{\pi_{\mathcal{C}}} v^{\pi_{\mathcal{C}}}(\mathcal{C}) + \max_{\pi_{\mathcal{D}}} v^{\pi_{\mathcal{D}}}(\mathcal{D}), \forall \mathcal{C}, \mathcal{D} \subset \mathcal{N}, \mathcal{C} \cap \mathcal{D} = \phi$ holds. In this paper, we denote the joint policy of the whole agents as $\pi = \times_{i \in \mathcal{N}} \pi_{i}(a_{i}|s)$ and assume that each agent can observe the global state.
        
        \begin{lemma}[\citeauthor{shapley1971cores} \citeyear{shapley1971cores}; \citeauthor{chalkiadakis2011computational} \citeyear{chalkiadakis2011computational}]
            1) Every convex game has a non-empty core. 2) If a solution $(\mathcal{CS}, \textbf{x})$ is in the core of a characteristic function game $\langle \mathcal{N}, v \rangle$ and the payoff distribution scheme is efficient, then $v(\mathcal{CS}) \geq v(\mathcal{CS}')$ for every coalition structure $\mathcal{CS}' \in \mathcal{CS}_{\mathcal{N}}$.
            \label{lem: cfs_core->v}
        \end{lemma}
        
        \begin{theorem}
            With the efficient payoff distribution scheme, for an extended convex game (ECG), one solution in the core must exist with the grand coalition and the objective is $\max_{\pi} v^{\pi}(\{\mathcal{N}\})$, which can lead to the maximal social welfare, i.e., $\max_{\pi} v^{\pi}(\{\mathcal{N}\}) \geq \max_{\pi} v^{\pi}(\mathcal{CS}')$ for every coalition structure $\mathcal{CS}' \in \mathcal{CS}_{\mathcal{N}}$.
            \begin{proof}
                This theorem is proved based on Lemma \ref{lem: cfs_core->v}. See Appendix for the complete proof.
            \end{proof}
            \label{the: CFS objective}
        \end{theorem}
        
        \begin{corollary}
            For an extended convex game (ECG) with the grand coalition, Shapley value must be in the core.
            \begin{proof}
                Since Shapley value must be in the core for a CG with the grand coalition \cite{shapley1971cores} and ECG still conserves the property of CG, this statement holds.
            \end{proof}
            \label{coro: shapley core}
        \end{corollary}
        
    % \subsection{Comparing ECG with Global Reward Game}
    % \label{subsec:compare_esg_global}
        As seen from Theorem \ref{the: CFS objective}, with an appropriate efficient payoff distribution scheme, an ECG with the grand coalition is actually equivalent to a global reward game. Both of them aim to maximize the global value (i.e., the global reward). Here, we assume that the agents in a global reward game are regarded as the grand coalition. Consequently, the local reward approach in ECG is feasible in the global reward game.
        
    \subsection{Looking into the Shared Reward Approach by the View of ECG}
    \label{subsec:fully}
\begin{figure}
            \centering
            \includegraphics*[scale=0.35]{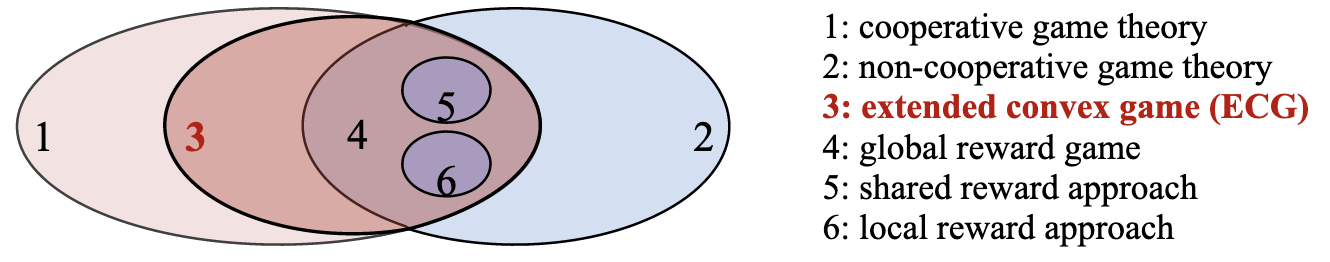}
            \caption{The relationship between the concepts mentioned in this paper.}
            \label{fig:venn_relation}
        \end{figure}

        Shared reward approach assigns each agent the global reward directly in the global reward game \cite{panait2005cooperative}. Each agent unilaterally maximizes this global reward to seek its optimal policy such that
        \begin{equation}
            \max_{\pi_{i}} v^{\pi}(\{\mathcal{N}\}) = \max_{\pi_{i}} \mathbb{E}_{\pi}[ \ \sum_{t=1}^{\infty} r_{t}(\mathcal{N}) \ ],
            \label{eq:full_equal}
        \end{equation}
        where $r_{t}(\mathcal{N})$ is the global reward and $\pi = \times_{i \in \mathcal{N}} \pi_{i}$. If $v^{\pi}(\{\mathcal{N}\})$ is multiplied by a normalizing factor, i.e., $1/|\mathcal{N}|$, then the objective of the new optimization problem for each agent $i$ is equivalent to Eq.\ref{eq:full_equal}. We can express it mathematically as
        \begin{equation}
            \begin{split}
                \max_{\pi_{i}} v^{\pi}(\{\mathcal{N}\}) = \max_{\pi_{i}} \sum_{i \in \mathcal{N}} \text{x}_{i} 
                &= \max_{\pi_{i}} \text{x}_{i} \\
                &= \max_{\pi_{i}} \frac{1}{|\mathcal{N}|} \cdot v^{\pi}(\{\mathcal{N}\}).
            \end{split}
        \end{equation}
        Then, the credit assigned to each agent in the shared reward approach is actually $x_{i} = v^{\pi}(\{\mathcal{N}\})/|\mathcal{N}|$, and the sum of the whole agents' credits is equal to the global reward. It suffices the condition of efficient payoff distribution scheme. Therefore, we show that the shared reward approach is an efficient payoff distribution scheme in an ECG with the grand coalition (i.e., a global reward game). Nevertheless, from the view of ECG, the shared reward approach is not guaranteed to find the optimal solution. By Corollary \ref{coro: shapley core} and Theorem \ref{the: CFS objective}, we know that Shapley value has to lie in the core and theoretically guarantees the convergence to the maximal global value. This is one of the reasons why we are interested in this local reward approach. To clarify the concepts we mentioned before, we draw a Venn diagram shown as Fig.\ref{fig:venn_relation}.
        
    \subsection{Shapley Q-value}
    \label{subsec:formulate the shapley value in marl}
         Although the shared reward approach successfully solves out a global reward game in practice, it has been shown that local reward approach gives the faster convergence rate \cite{balch1997learning,balch1999reward}. For the two aforementioned reasons, we use the Shapley value (i.e., a local reward approach) for the credit assignment to each agent. Because $v^{\pi_{\mathcal{C}'}}(\mathcal{C}')$ represents the global reward earned by coalition $\mathcal{C}'$ in an ECG, we can model it as a Q-value, where $s$ represents the state and $\textbf{a}_{\mathcal{C}'} = (a_{i})_{i \in \mathcal{C}'}$. According to Eq.\ref{eq: shapley_value_org}, the Shapley Q-value of each agent can be written as 
        \begin{align}
            &\Phi_{i}(\mathcal{C}) = Q^{\pi_{\mathcal{C} \cup \{i\}}}(s, \textbf{a}_{\mathcal{C} \cup \{i\}}) - Q^{\pi_{\mathcal{C}}}(s, \textbf{a}_{C}), \label{eq:shapley_q_value_1}\\
            &Q^{\Phi_{i}}(s, a_{i}) = \sum_{\mathcal{C} \subseteq \mathcal{N} \backslash \{i\}} \frac{|\mathcal{C}|!(|\mathcal{N}|-|\mathcal{C}|-1)!}{|\mathcal{N}|!} \cdot \Phi_{i}(\mathcal{C}). \label{eq:shapley_q_value_2}
        \end{align}
        
    \subsubsection{Approximate Marginal Contribution}
    \label{subsec:approximate marginal contribution}
        As seen from Eq.\ref{eq:shapley_q_value_1}, it is difficult and unstable to learn two Q-value functions (where one is for representing the Q-value of $\mathcal{C} \cup \{i\}$ and the other is for representing the Q-value of $\mathcal{C}$) to estimate the marginal contributions (i.e., making the difference between two Q-values) for different coalitions. To mitigate this problem, we propose a method called approximate marginal contribution (AMC) to directly estimate the marginal contribution of each coalition (i.e., $\Phi_{i}(\mathcal{C})$).
        
        In cooperative game theory, each agent is assumed to join in the grand coalition sequentially. $|\mathcal{C}|!(|\mathcal{N}|-|\mathcal{C}|-1)! / |\mathcal{N}|!$ in Eq.\ref{eq: shapley_value_org}, denoted as $Pr(\mathcal{C}|\mathcal{N} \backslash \{i\})$, is interpreted as that an agent randomly joins in an existing coalition $\mathcal{C}$ (which could be empty) to form the complete grand coalition with the subsequent agents \cite{chalkiadakis2011computational}. According to this interpretation, we model a function to approximate the marginal contribution directly such that
        \begin{equation}
            \hat{\Phi}_{i}(s, \textbf{a}_{\mathcal{C} \cup \{i\}}): \mathcal{S} \times \mathcal{A}_{\mathcal{C} \cup \{i\}} \mapsto \mathbb{R},
            \label{eq:amc_org}
        \end{equation}
        where $\mathcal{S}$ is the state space; $\mathcal{C}$ is the ordered coalition that agent $i$ would like to join in; $\mathcal{A}_{\mathcal{C} \cup \{i\}} = (\mathcal{A}_{j})_{j \in \mathcal{C} \cup \{i\}}$, and the actions are ordered. For example, if the order of a coalition is $(0,  2)$ (where $0$ and $2$ are two agents), then $\textbf{a}_{\mathcal{C} \cup \{1\}} = (a_{0}, a_{2}, a_{1})$. By such a formulation, we believe that the property of the marginal contribution (i.e., mapping from every possible combination of coalition $\mathcal{C}$ and agent $i$ to a numerical value) can be maintained. Hence, AMC is reasonable to replace the exact marginal contribution. In practice, we represent $\textbf{a}_{\mathcal{C} \cup \{i\}}$ by the concatenation of each agent's action vector. To keep the input size of $\hat{\Phi}_{i}(s, \textbf{a}_{\mathcal{C} \cup \{i\}})$ constant in different cases, we fix the actions as the concatenation of all agents' actions and mask the actions of irrelevant agents (i.e., the agents who do not exist in the coalition) with zeros. 
        
    \subsubsection{Approximate Shapley Q-value}
    \label{subsec:approximate shapley q-value}
        Followed by the interpretation above, Shapley Q-value can be rewritten as 
        \begin{equation}
            Q^{\Phi_{i}}(s, a_{i}) = \mathbb{E}_{\mathcal{C} \sim Pr(\mathcal{C}|\mathcal{N} \backslash \{i\})}[ \ \Phi_{i}(\mathcal{C}) \ ].
            \label{eq:asq_org}
        \end{equation}
        To enable Eq.\ref{eq:asq_org} to be tractable in realization, we can sample $Q^{\Phi_{i}}(s, a_{i})$ here. Also, substituting AMC in Eq.\ref{eq:amc_org} for the marginal contribution in Eq.\ref{eq:shapley_q_value_1}, we can approximate Shapley Q-value such that 
        \begin{equation}
            Q^{\Phi_{i}}(s, a_{i}) \approx \frac{1}{M} \sum_{k=1}^{M} \hat{\Phi}_{i}(s, \textbf{a}_{\mathcal{C}_{k} \cup \{i\}}), \ \ \forall \mathcal{C}_{k} \sim Pr(\mathcal{C}|\mathcal{N} \backslash \{i\}).
            \label{eq:shapley_approx}
        \end{equation}
        
    \subsection{Shapley Q-value Deep Deterministic Policy Gradient}
    \label{subsec:Shapley Q-value deep deterministic policy gradient}
        In an ECG with the grand coalition, each agent only needs to maximize its own Shapley value (i.e., $Q^{\Phi_{i}}(s, a_{i})$) so that $\max_{\pi} v^{\pi}(\mathcal{N})$ can be achieved such that
        \begin{equation}
            \begin{split}
                \max_{\pi} v^{\pi}(\mathcal{N}) = \max_{\pi} \sum_{i \in \mathcal{N}} \text{x}_{i} 
                &= \sum_{i \in \mathcal{N}} \max_{\pi_{i}} \text{x}_{i} \\
                &= \sum_{i \in \mathcal{N}} \max_{\pi_{i}} Q^{\Phi_{i}}(s, a_{i}).
            \end{split}
        \end{equation}
        Therefore, if we can show that $\max_{\pi_{i}} Q^{\Phi_{i}}(s, a_{i})$ for each agent $i$ is approached, then we show that the maximal global value $\max_{\pi} v^{\pi}(\mathcal{N})$ is sufficient to be achieved. Now, the problem transfers to solving $\max_{\pi_{i}} Q^{\Phi_{i}}(s, a_{i})$ for each agent $i$. Aforementioned, a global reward game is identical to a potential game \footnote{A potential game is a game where there exists a potential function \cite{monderer1996potential}.}. Additionally, \newciteauthor{monderer1996potential} showed that in a potential game there exists a pure Nash equilibrium (i.e., a deterministic optimal policy solution). For these reasons, we apply deterministic policy gradient (DPG) \cite{silver2014deterministic} to search out a deterministic optimal policy.
        If we substitute Shapley Q-value for $Q_{i}^{\pi}(s, a_{i})$ in DPG, we can directly write the policy gradient of each agent $i$ such that
        \begin{equation}
            \nabla_{\theta_{i}}J(\theta_{i}) = \mathbb{E}_{s \sim \rho^{\mu}}[\nabla_{\theta_{i}} \mu_{\theta_{i}}(s) \nabla_{a_{i}} Q^{\Phi_{i}}(s, a_{i})|_{a_{i}=\mu_{\theta_{i}}(s)}],
            \label{eq:shapley_q_value_pg}
        \end{equation}
        where $Q^{\Phi_{i}}(s, a_{i})$ is the Shapley Q-value for agent $i$ and $\mu_{\theta_{i}}$ is agent $i$'s deterministic policy, parameterized by $\theta_{i}$. A global reward is received each time step in a global reward game. Since the Shapley Q-value of each agent is correlated to a local reward, we cannot update each $\hat{Q}^{\Phi_{i}}(s, a_{i})$ directly by the global reward. However, benefited by the property of efficiency (see Section \ref{subsec:shapley value}), we can solve it out according to the minimization problem such that
        \begin{equation}
            \begin{split}
                \min_{\omega_{1}, \omega_{2}, ..., \omega_{|\mathcal{N}|}} &\mathbb{E}_{s^{t}, \textbf{a}_{\mathcal{N}}^{t}, r^{t}(\mathcal{N}), s^{t+1}}[ \ \ \frac{1}{2}( \ \  r^{t}( \mathcal{N}) \\
                &+ \gamma \sum_{i \in \mathcal{N}} Q^{\Phi_{i}}(s^{t+1}, a_{i}^{t+1}; \omega_{i})|_{a_{i}^{t+1}=\mu_{\theta_{i}}(s^{t+1})} \\
                &- \sum_{i \in \mathcal{N}} Q^{\Phi_{i}}(s^{t}, a_{i}^{t}; \omega_{i}) \ \ )^{2} \ \ ],
            \end{split}
            \label{eq:shapley_q_value_critic}
        \end{equation}
        where $r(\mathcal{N})$ is the global reward received from the environment each time step and $Q^{\Phi_{i}}(s, a_{i}; \omega_{i})$ for each agent $i$ is parameterized by $\omega_{i}$. Constrained by this objective function, Shapley Q-value suffices the property of efficiency. Accordingly, the condition of efficient payoff distribution scheme stated in Theorem \ref{the: CFS objective} is promised. Because Shapley Q-value takes all of feasible agents' actions and states as input, our algorithm actually uses the centralized critic. Nevertheless, the policies are decentralized in execution.

        \newciteauthor{silver2014deterministic} showed that DPG has the familiar machinery of policy gradient. Besides, \newciteauthor{sutton2018reinforcement} emphasized that with a small learning rate, policy gradient algorithm can converge to a local optimum. Consequently, we can conclude that with a small learning rate, each agent can find a local maximizer and the global value $v^{\pi}(\mathcal{N})$ converges to a local maximum. The convexity of $v^{\pi}(\mathcal{N})$ is impossible to be guaranteed in applications, so the global maximum stated in Theorem \ref{the: CFS objective} may not be always fulfilled. 
            
    \subsubsection{Implementation}
    \label{subsec:implementation}
        In implementation, we replace the exact Shapley Q-value by the approximate Shapley Q-value in Eq.\ref{eq:shapley_approx}. Accordingly, the Shapley Q-value here is actually a linear combination of the approximate marginal contributions (AMCs) (i.e., $\hat{\Phi}_{i}(s, \textbf{a}_{\mathcal{C} \cup \{i\}}; \omega_{i})$). Thanks to the powerful off-policy training strategy and function approximation on AMCs by the deep neural networks, we use the deep deterministic policy gradient (DDPG) method \cite{lillicrap2015continuous} for learning policies. Additionally, we apply the reparameterization technique called Gumbel-Softmax trick \cite{DBLP:conf_iclr_JangGP17} to deal with the discrete action space. Since our algorithm aims to find the optimal policies by Shapley Q-value and DDPG, we call it Shapley Q-value deep deterministic policy gradient (SQDDPG). The pseudo code for the SQDDPG is given in Appendix.
        
%===============================================================================

\section{Experiments}
\label{sec:result}
    We evaluate SQDDPG on Cooperative Navigation, Prey-and-Predator \cite{lowe2017multi} and Traffic Junction \cite{NIPS2016_6398} \footnote{The code of experiments is available on: \url{https://github.com/hsvgbkhgbv/SQDDPG}}. In the experiments, we compare SQDDPG with two Independent algorithms (with decentralised critics), e.g., Independent DDPG (IDDPG) \cite{lillicrap2015continuous} and Independent A2C (IA2C) \cite{sutton2018reinforcement}, and two state-of-the-art methods with centralised critics, e.g., MADDPG \cite{lowe2017multi} and COMA \cite{foerster2018counterfactual}. In the experiments, to keep the fairness of comparison, the policy and critic networks for all MARL algorithms are parameterized by MLPs. All models are trained by the Adam optimizer \cite{kingma2014adam}. The details of experimental setups are given in Appendix.
    
    \subsection{Cooperative Navigation}
    \label{subsec:cooperative_navigation}
        \subsubsection{Environment Settings} In this task, there are 3 agents and 3 targets. Each agent aims to move to a target, with no prior allocations of the targets to each agent. The state of each agent in this environment includes its current position and velocity, the displacement to the three targets, and the displacement to other agents. The action space of each agent is \texttt{move\_up}, \texttt{move\_down}, \texttt{move\_right}, \texttt{move\_left} and \texttt{stay}. The global reward of this environment is the negative sum of the distance between each target and the nearest agent to it. Besides, if a collision happens, then the global reward will be reduced by 1.
        
        \subsubsection{Results} 
\begin{figure}
            \centering
            \includegraphics[scale=0.3]{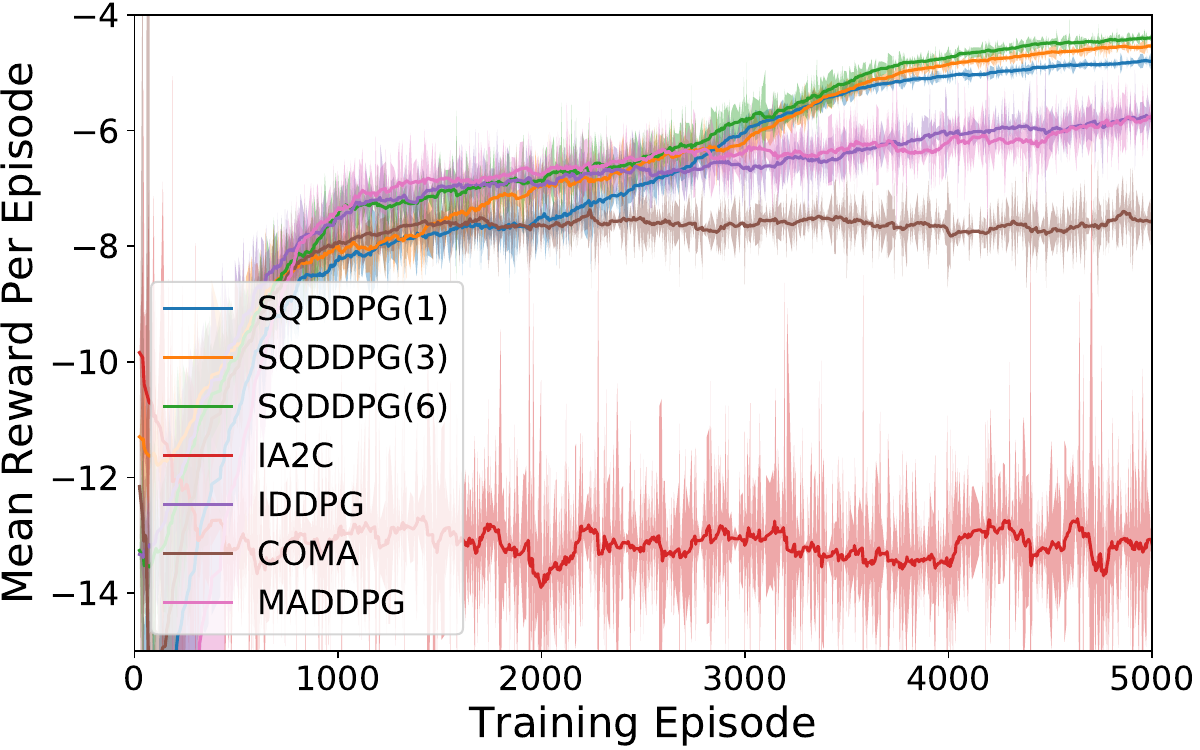}
            \caption{Mean reward per episode during training in Cooperative Navigation. SQDDPG(n) indicates SQDDPG with the sample size (i.e., M in Eq.\ref{eq:shapley_approx}) of n. In the rest of experiments, since only SQDDPG with the sample size of 1 is run, we just use SQDDPG to represent SQDDPG(1).}
            \label{fig:3_agents_reward_spread}
        \end{figure}

As seen from Fig.\ref{fig:3_agents_reward_spread}, SQDDPGs with variant sample sizes (i.e., M in Eq.\ref{eq:shapley_approx}) outperform the baselines on the convergence rate. We believe that if more training episodes are permitted, the algorithms except for IA2C can achieve the similar performance as that of SQDDPG. Therefore, our result supports the previous argument that the local reward approach converges faster than the global reward approach \cite{balch1997learning,balch1999reward}. As the sample size grows, the approximate Shapley Q-value estimation in Eq.\ref{eq:asq_org} could be more accurate and easier to converge to the optimal value. This explains the reason why the convergence rate of SQDDPG becomes faster when the sample size increases. Since we show that SQDDPG with the sample size of 1 can finally obtain nearly the same performance as that of other variants, we just run it in the rest of experiments to reduce the computational complexity.
        
    \subsection{Prey-and-Predator}
    \label{subsec:prey_predator}
        \subsubsection{Environment Settings} In this task, we can only control three predators and the prey is a random agent. The aim of each predator is coordinating to capture the prey by as less steps as possible. The state of each predator contains its current position and velocity, the respective displacement to the prey and other predators, and the velocity of the prey. The action space is the same as that defined in Cooperative Navigation. The global reward is the negative minimal distance between any predator and the prey. If the prey is caught by any predator, the global reward will be added by 10 and the game terminates.
        
        \begin{figure}
            \centering
            \includegraphics[scale=0.3]{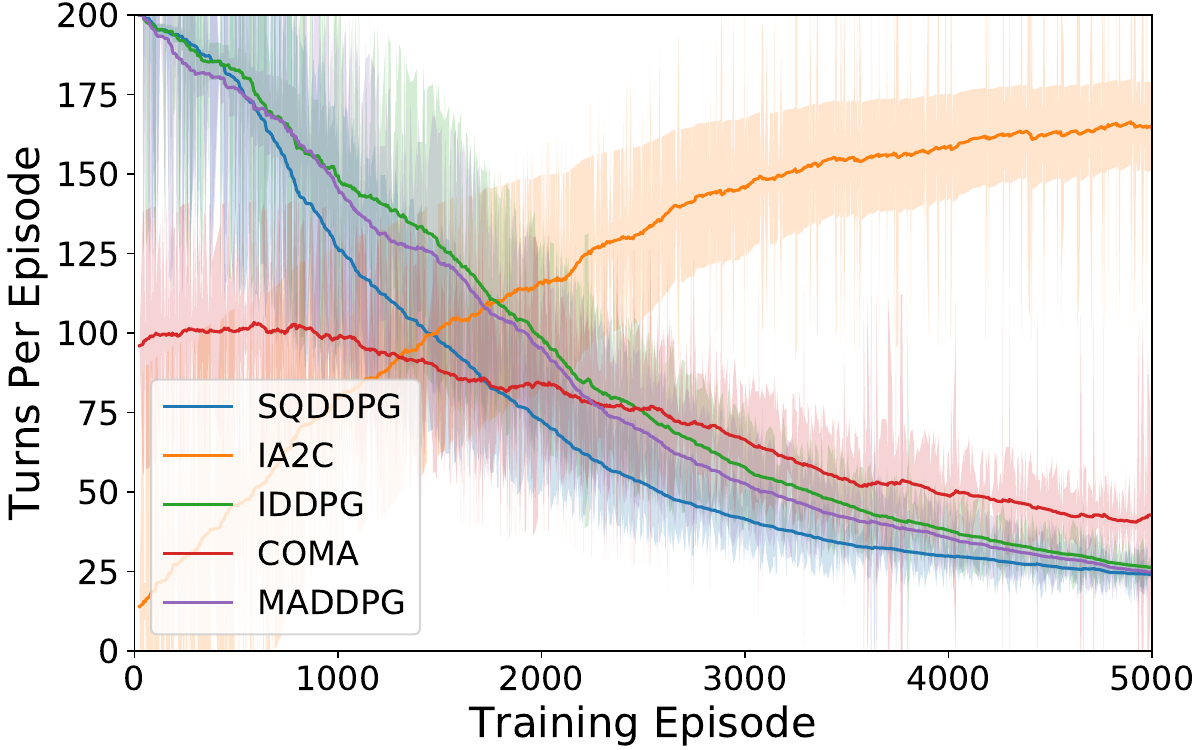}
            \caption{Turns to capture the prey per episode during training in Prey-and-Predator.}
            \label{fig:3_agents_turns_tag}
        \end{figure}
        
        \subsubsection{Results} 
\begin{figure*}
            \centering
            \includegraphics*[scale=0.3]{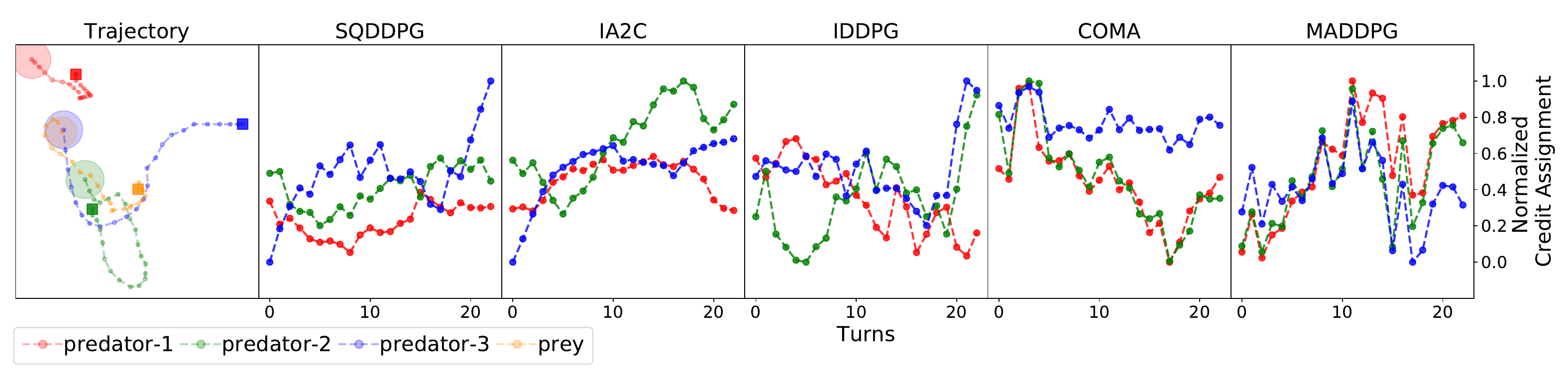}
            \caption{Credit assignment to each predator for a fixed trajectory. The leftmost figure records a trajectory sampled by an expert policy. The square represents the initial position whereas the circle indicates the final position of each agent. The dots on the trajectory indicates each agent's temporary positions. The other figures show the normalized credit assignments generated by different MARL algorithms according to this trajectory.}
            \label{fig:credit_vis}
        \end{figure*}

As Fig.\ref{fig:3_agents_turns_tag} shows, SQDDPG converges fastest with around 25 turns to capture the prey finally, followed by MADDPG and IDDPG. To study and understand the credit assignment, we visualize the Q-values of each MARL algorithm for one randomly selected trajectory of states and actions from an expert policy. For the convenient visualization, we normalize the Q-values by min-max normalization \cite{patro2015normalization} for each MARL algorithm. We can see from Fig.\ref{fig:credit_vis} that the credit assignment of SQDDPG is more explainable than that of the baselines. Specifically, it is intuitive that the credit assigned to each agent by SQDDPG is inversely proportional to its distance to the prey. However, other MARL algorithms do not explicitly show such a property. To validate this hypothesis, we also evaluate it quantitatively by Pearson correlation coefficient (PCC) on 1000 randomly selected transition samples for the correlation between the credit assignment and the reciprocal of each predator's distance to the prey. The value of PCC is greater, and the inverse proportion is stronger. As Tab.\ref{tab:correlation} shows, SQDDPG expresses the inverse proportion significantly, with the PCC of 0.3210 and the two-tailed p-value of 1.9542e-19. If a predator is closer to the prey, it is more likely to catch the prey and the contribution of that predator should be more significant. Consequently, we demonstrate that the credit assignment of Shapley Q-value is fair.
        
        \begin{table*}[ht!]
            \centering
            \begin{tabular}{cccccc}
                \toprule
                    \textbf{} & \textbf{IA2C} & \textbf{IDDPG} & \textbf{COMA} & \textbf{MADDPG} & \textbf{SQDDPG}  \\ 
                \midrule 
                    coefficient        &  0.0508  & 0.0061 & 0.1274 & 0.0094 & \textbf{0.3210} \\
                    two-tailed p-value &  1.6419e-1  & 8.6659e-1 & 4.6896e-4 & 7.9623e-1 & \textbf{1.9542e-19} \\
            \bottomrule
            \end{tabular}
            \caption{The Pearson correlation coefficient between the credit assignment to each predator and the reciprocal of its distance to the prey. This test is conducted by 1000 randomly selected episode samples.}
            \label{tab:correlation}
        \end{table*}
        
        \begin{table*}[ht!]
        \centering
        \begin{tabular}{cccccc}
            \toprule
                \textbf{Difficulty} & \textbf{IA2C} & \textbf{IDDPG} & \textbf{COMA} & \textbf{MADDPG} & \textbf{SQDDPG}  \\ 
            \midrule 
                Easy & 65.01\%   & 93.08\%  & 93.01\%   & \textbf{93.72\%}   & 93.26\%  \\
                Medium  & 67.51\%  & 84.16\% & 82.48\%  & 87.92\%         & \textbf{88.98\%} \\
                Hard  & 60.89\% & 64.99\%  & 85.33\%   & 84.21\%         & \textbf{87.04\%} \\ 
        \bottomrule
        \end{tabular}
        \caption{The success rate on Traffic Junction, tested with 20, 40, and 60 steps per episode in easy, medium and hard versions respectively. The results are obtained by running each algorithm after training for 1000 episodes.}
        \label{tab:traffic_junction}
        \end{table*}
        
    \subsection{Traffic Junction}
    \label{subsec:traffic_junction}
        \subsubsection{Environment Settings} In this task, cars move along the predefined routes which intersect on one or more traffic junctions. At each time step, new cars are added to the environment with the probability $p_{\text{arrive}}$, and the total number of cars is limited below $N_{\max}$. After a car finishes its mission, it will be removed from the environment and possibly sampled back to a new route. Each car has a limited vision of 1, which means it can only observe the circumstance within the 3x3 region surrounding it. No communication between cars is permitted in our experiment, in contrast to the others' experiments on the same task \cite{NIPS2016_6398,das2018tarmac}. The action space of each car is \texttt{gas} and \texttt{brake}, and the global reward function is $\sum_{i=1}^{N} \text{- }0.01t_{i}$, where $t_{i}$ is the time steps that car $i$ is continuously active on the road in one mission and $N$ is the total number of cars. Additionally, if a collision occurs, the global reward will be reduced by 10. We evaluate the performance by the success rate (i.e., the episode that no collisions happen).
        
        \subsubsection{Results} We compare our method with the baselines on the easy, medium and hard version of Traffic Junction. The easy version is constituted of one traffic junction of two one-way roads on a $7\times7$ grid with $N_{\max}\text{ = }5$ and $p_{\text{arrive}}\text{ = }0.3$. The medium version is constituted of one traffic junction of two-way roads on a $14\times14$ grid with $N_{\max}\text{ = }10$ and $p_{\text{arrive}}\text{ = }0.2$. The hard version is constituted of four connected traffic junctions of two-way roads on a $18\times18$ grid with $N_{\max}\text{ = }20$ and $p_{\text{arrive}}\text{ = }0.05$. From Tab.\ref{tab:traffic_junction}, we can see that on the easy version, except for IA2C, other algorithms can get a success rate over $93\%$, since this scenario is too easy. On the medium and hard version, SQDDPG outperforms the baselines with the success rate of $88.98\%$ on the medium version and $87.04\%$ on the hard version. Moreover, the performance of SQDDPG significantly exceeds that of no-communication algorithms reported as $84.9\%$ and $74.1\%$ in \cite{das2018tarmac}. We demonstrate that SQDDPG can solve out the large-scale problems.
    
    \subsection{Discussion}
    \label{subsec:discussion}
        In the experimental results, it is surprising that IDDPG achieves a good performance. The possible reason could be that a potential game (i.e., a global reward game) can be solved by the fictitious play \cite{monderer1996potential} and DDPG is analogous to it, finding an optimal deterministic policy by fitting the other agents' behaviours. However, the convergence rate is not guaranteed when the number of agents becomes large, such as the result shown in the hard version of Traffic Junction. To deal with both of competitive and cooperative games, MADDPG assigns each agent a centralised critic to estimate the global reward. Theoretically the credits assigned to agents are identical, though in experiment it does not always exhibit this property. The possible reason could be the bias existing in the Q-values. We can see from Tab.\ref{tab:correlation} that both of COMA and SQDDPG exhibit the feature of credit assignment. Nonetheless, SQDDPG performs more significantly on the fairness. This well validates our motivation of importing the local reward approach (e.g., Shapley value) to solve out the credit assignment problems.
        
%===============================================================================

\section{Conclusion}
\label{sec:conclusion}
    We introduce the cooperative game theory to extend the existing global reward game to a broader framework called extended convex game (ECG). Under this framework, we propose an algorithm namely Shapley Q-value deep deterministic policy gradient (SQDDPG), leveraging a local reward approach called Shapley Q-value, which is theoretically guaranteed to find out the optimal solution in an ECG with the grand coalition (i.e., a global reward game). We evaluate SQDDPG on three global reward games (i.e., Cooperative Navigation, Prey-and-Predator and Traffic Junction) and show the promising performance compared with the baselines (e.g., MADDPG, COMA, IDDPG and IA2C). During the experiments, SQDDPG exhibits the fair credit assignments and fast convergence rate. In the future work, we plan to dynamically group the agents at each time step with some theoretical guarantees and jump out of the restriction of global reward game.

%===============================================================================

\section*{Acknowledgements}
This work is sponsored by EPSRC-UKRI Innovation Fellowship EP/S000909/1. Jianhong Wang especially thanks Ms Yunlu Li for useful and patient explanations on mathematics. 

%===============================================================================

\newpage
\section*{Appendix}
\appendix

    \subsection*{Algorithm}
        \label{appendix:algo}
        In this section, we show the pseudo code of Shapley Q-value deep deterministic policy gradient (SQDDPG) in Algorithm \ref{alg:sqpg}.
        
        \begin{algorithm*}[ht!]
        \caption{Shapley Q-value deep deterministic policy gradient (SQDDPG)}
        \label{alg:sqpg}
            \begin{algorithmic}[1]
                \State Initialize actor parameters $\theta_{i}$, and critic (AMC) parameters $\omega_{i}$ for each agent $i \in \mathcal{N}$
                \State Initialize target actor parameters $\theta_{i}'$, and target critic parameters $\omega_{i}'$ for each agent $i \in \mathcal{N}$
                \State Initialize the sample size $M$ for approximating Shapley Q-value
                \State Initialize the learning rate $\tau$ for updating target network
                \State Initialize the discount rate $\gamma$
                \For{episode = 1 to D}
                    \State Observe initial state $s_{1}$ from the environment
                    \For{t = 1 to T}
                        \State $u_{i} \gets \mu_{\theta_{i}}(s_{t}) + N_{t}$ for each agent $i$ \Comment{\small{Select action according to the current policy and exploration noise.}}
                        \normalsize
                        \State Execute actions $\textbf{u}_{t}=\times_{i \in \mathcal{N}} u_{i}$ and observe the global reward $r_{t}$ and the next state $s_{t+1}$
                        \State Store $(s_{t}, \textbf{u}_{t}, r_{t}, s_{t+1})$ in the replay buffer $\mathcal{B}$
                        \State Sample a minibatch of G samples $(s_{k}, \textbf{u}_{k}, r_{k}, s_{k+1})$ from $\mathcal{B}$
                        \State Get $\textbf{a}_{k} = \times_{i \in \mathcal{N}} \mu_{\theta_{i}}(s_{k})$ for each sample $(s_{k}, \textbf{u}_{k}, r_{k}, s_{k+1})$
                        \State Get $\hat{\textbf{a}}_{k} = \times_{i \in \mathcal{N}} \mu_{\theta_{i}'}(s_{k+1})$ for each sample $(s_{k}, \textbf{u}_{k}, r_{k}, s_{k+1})$
                        \For{each agent $i$} \Comment{\small{This procedure can be implemented in parallel.}}
                            \normalsize
                            \State Sample $M$ ordered coalitions by $\mathcal{C} \sim Pr(\mathcal{C} | \mathcal{N} \backslash \{i\})$
                            \For{each sampled coalition $\mathcal{C}_{m}$} \Comment{\small{This procedure can be implemented in parallel.}}
                                \normalsize
                                \State Order each $\textbf{a}_{k}$ by $\mathcal{C}_{m}$ and mask the irrelevant agents' actions, storing it to $\textbf{a}_{k}^{m}$
                                \State Order each $\hat{\textbf{a}}_{k}$ by $\mathcal{C}_{m}$ and mask the irrelevant agents' actions, storing it to $\hat{\textbf{a}}_{k}^{m}$
                                \State Order each $\textbf{u}_{k}$ by $\mathcal{C}_{m}$ and mask the irrelevant agents' actions, storing it to $\textbf{u}_{k}^{m}$
                            \EndFor
                            \State Get $Q_{k}^{\Phi_{i}}(s_{k}, a_{i}; \omega_{i}) \gets \frac{1}{M} \sum_{m=1}^{M} \hat{\Phi}_{i}(s_{k}, \textbf{a}_{k}^{m}; \omega_{i})$ for each sample $(s_{k}, \textbf{u}_{k}, r_{k}, s_{k+1})$
                            \State Get $Q_{k}^{\Phi_{i}}(s_{k}, u_{i}; \omega_{i}) \gets \frac{1}{M} \sum_{m=1}^{M} \hat{\Phi}_{i}(s_{k}, \textbf{u}_{k}^{m}; \omega_{i})$ for each sample $(s_{k}, \textbf{u}_{k}, r_{k}, s_{k+1})$
                            \State Get $Q_{k}^{\Phi_{i}}(s_{k}, \hat{a}_{i}; \omega_{i}') \gets \frac{1}{M} \sum_{m=1}^{M} \hat{\Phi}_{i}(s_{k}, \hat{\textbf{a}}_{k}^{m}; \omega_{i}')$ for each sample $(s_{k}, \textbf{u}_{k}, r_{k}, s_{k+1})$
                            \normalsize
                            \State Update $\theta_{i}$ by deterministic policy gradient according to Eq.\ref{eq:shapley_q_value_pg}:
                            \begin{equation*}
                                \nabla_{\theta_{i}}J(\theta_{i}) = \frac{1}{G} \sum_{k=1}^{G} \nabla_{\theta_{i}} \mu_{\theta_{i}}(s_{k}) \nabla_{a_{i}} Q_{k}^{\Phi_{i}}(s_{k}, a_{i}; \omega_{i})|_{a_{i}=\mu_{\theta_{i}}(s_{k})}
                            \end{equation*}
                        \EndFor
                        \State Set $y_{k} = r_{k} + \gamma \sum_{i \in \mathcal{N}} Q_{k}^{\Phi_{i}}(s_{k}, \hat{a}_{i}; \omega_{i}')$ for each sample $(s_{k}, \textbf{u}_{k}, r_{k}, s_{k+1})$
                        \State Update $\omega_{i}$ for each agent $i$ by minimizing the optimization problem according to Eq.\ref{eq:shapley_q_value_critic}:
                        \begin{equation*}
                            \min_{\omega_{i}} \frac{1}{G} \sum_{k=1}^{G} \frac{1}{2}( \ y_{k} - \sum_{i \in \mathcal{N}} Q_{k}^{\Phi_{i}}(s_{k}, u_{i}; \omega_{i}) \ )^{2}
                        \end{equation*}
                        \State Update target network parameters for each agent $i$:
                            \begin{align*}
                                &\theta_{i}' \gets \tau \theta_{i} + (1 - \tau) \theta_{i}'\\
                                &\omega_{i}' \gets \tau \omega_{i} + (1 - \tau) \omega_{i}'
                            \end{align*}
                    \EndFor
                \EndFor
            \end{algorithmic}
        \end{algorithm*}
        
    \subsection*{Proof of Theorem 1}
        \label{appendix:cfs}
        \begingroup
            \def\thetheorem{\ref{the: CFS objective}}
                \begin{theorem}
                    With the efficient payoff distribution scheme, for an extended convex game (ECG), one solution in the core can be certainly found with the grand coalition and the objective is $\max_{\pi} v^{\pi}(\{\mathcal{N}\})$, which can lead to the maximal social welfare, i.e., $\max_{\pi} v^{\pi}(\{\mathcal{N}\}) \geq \max_{\pi} v^{\pi}(\mathcal{CS}')$ for every coalition structure $\mathcal{CS}' \in \mathcal{CS}_{\mathcal{N}}$.
                    \begin{proof}
                        The proof is as follows.
                        
                        As we defined before, in an ECG, after allocating coalitions each agent will further maximize the cumulative rewards of his coalition by the optimal policy. Now, we denote the optimal value of an arbitrary coalition $\mathcal{C}'$ as $v^{*}(\mathcal{C}') = \max_{\pi_{\mathcal{C}'}} v^{\pi_{\mathcal{C}'}}(\mathcal{C}')$, where $\pi_{\mathcal{C}'} = \times_{i \in \mathcal{C}'} \pi_{i}$. Similarly, we can define the optimal value given an arbitrary coalition structure $\mathcal{CS}$ as $v^{*}(\mathcal{CS})=\sum_{\mathcal{C} \in \mathcal{CS}} v^{*}(\mathcal{C})$. If we re-write the value function defined above, ECG can be reformulated to CG. For this reason, we can directly use the results in Lemma \ref{lem: cfs_core->v} here to complete the proof.
                        
                        First, we aim to show that \textbf{(i)} \textit{In an ECG, there exists an efficient payoff distribution scheme $\textbf{x}$ that $v^{*}(\{\mathcal{N}\}) \geq v^{*}(\mathcal{CS}')$ for any $\mathcal{CS}' \in \mathcal{CS}_{\mathcal{N}}$, and $(\{\mathcal{N}\}, \textbf{x})$ is a solution in the core.}
                        
                        Suppose for the sake of contradiction that $(\{\mathcal{N}\}, \textbf{x})$ is not in the core, but due to Statement (1) in Lemma \ref{lem: cfs_core->v}, there must exist a coalition structure $\mathcal{CS}$ other than $\{\mathcal{N}\}$ satisfies the result that $(\mathcal{CS}, \hat{\textbf{x}})$ is in the core. According to Statement (2) in Lemma \ref{lem: cfs_core->v}, since CG is a subclass of characteristic function games, with an efficient payoff distribution scheme we can get that 
                        \begin{equation}
                            v^{*}(\mathcal{CS}) = \sum_{\mathcal{C} \in \mathcal{CS}} v^{*}(\mathcal{C}) \geq v^{*}(\{\mathcal{N}\}).
                            \label{eq:contra1}
                        \end{equation}
                        
                        On the other hand, because of the property of ECG, i.e., 
                        \begin{equation}
                            \begin{split}
                                \max_{\pi_{\mathcal{C} \cup \mathcal{D}}} v^{\pi_{\mathcal{C} \cup \mathcal{D}}}(\mathcal{C} \cup \mathcal{D}) \geq \max_{\pi_{\mathcal{C}}} v^{\pi_{\mathcal{C}}}(\mathcal{C}) + \max_{\pi_{\mathcal{D}}} v^{\pi_{\mathcal{D}}}(\mathcal{D}), \\ \forall \mathcal{C}, \mathcal{D} \subset \mathcal{N}, \mathcal{C} \cap \mathcal{D} = \phi,
                            \end{split}
                        \end{equation}
                        
                        we have that 
                        \begin{align}
                            v^{*}(\{\mathcal{N}\}) &= \max_{\pi_{\mathcal{N}}} v^{\pi_{\mathcal{N}}}(\{\mathcal{N}\}) \notag\\
                            &= \max_{\pi_{\mathcal{N}}} v^{\pi_{\mathcal{N}}}(\mathcal{N}) \notag\\
                            &\geq \max_{\pi_{\mathcal{C}_{1}}} v^{\pi_{\mathcal{C}_{1}}}(\mathcal{C}_{1}) + \max_{\pi_{\mathcal{C}_{2}}} v^{\pi_{\mathcal{C}_{2}}}(\mathcal{C}_{2}) \ \ \ \ \notag \\ 
                            &(\forall \ \mathcal{C}_{1} \cup \mathcal{C}_{2} = \bigcup_{\mathcal{C} \in \mathcal{CS}} \mathcal{C}, \ \mathcal{C}_{1} \cap \mathcal{C}_{2} = \phi) \notag \\
                            &\geq \vdots \ \ \ \ \vdots \ \ \ \ \vdots \ \ \ \ \vdots \ \ \ \ \vdots \ \ \ \ \vdots \ \ \ \ \vdots \ \ \ \ \vdots \ \ \ \ \vdots \ \ \ \ \vdots \ \ \ \ \vdots \ \ \ \ \notag \\
                            &\text{(We further expand the terms similarly)} \notag\\
                            &\geq \sum_{\mathcal{C} \in \mathcal{CS}} \max_{\pi_{\mathcal{C}}} v^{\pi_{\mathcal{C}}}(\mathcal{C})\notag\\
                            &= v^{*}(\mathcal{CS})
                            \label{eq:contra2}
                        \end{align}
                        
                        By Eq.\ref{eq:contra1} and \ref{eq:contra2}, we can get that
                        \begin{equation}
                            v^{*}(\{\mathcal{N}\}) = v^{*}(\mathcal{CS}).
                            \label{eq:contra3}
                        \end{equation}
                        
                        According to the condition of efficient payoff distribution scheme, we can write:
                        \begin{equation}
                            v^{*}(\{\mathcal{N}\}) = \max_{\pi_{\mathcal{N}}} \sum_{i \in \mathcal{N}} \text{x}_{i} = \sum_{\mathcal{C} \in \mathcal{CS}} \max_{\pi_{\mathcal{C}}} \textbf{x}(\mathcal{C}),
                            \label{eq:contra4}
                        \end{equation}
                        \begin{equation}
                            v^{*}(\mathcal{CS}) = \sum_{\mathcal{C} \in \mathcal{CS}} v^{*}(\mathcal{C}) = \sum_{\mathcal{C} \in \mathcal{CS}} \max_{\pi_{\mathcal{C}}} \hat{\textbf{x}}(\mathcal{C}).
                            \label{eq:contra5}
                        \end{equation}
                        
                        By Eq.\ref{eq:contra3}, we can get that
                        \begin{equation}
                            \sum_{\mathcal{C} \in \mathcal{CS}} \max_{\pi_{\mathcal{C}}} \textbf{x}(\mathcal{C}) = \sum_{\mathcal{C} \in \mathcal{CS}} \max_{\pi_{\mathcal{C}}} \hat{\textbf{x}}(\mathcal{C}).
                            \label{eq:contra6}
                        \end{equation}
                        
                        By Eq.\ref{eq:contra6}, it is obvious that we can always find out a payoff distribution scheme $\textbf{x} = \hat{\textbf{x}}$ for the grand coalition $\{\mathcal{N}\}$.  Since $(\mathcal{CS}, \hat{\textbf{x}})$ is presumed to be in the core, $(\{\mathcal{N}\}, \hat{\textbf{x}})$ must satisfy the conditions of core. As a result, we derive that $(\{\mathcal{N}\}, \textbf{x})$ (where $\textbf{x} = \hat{\textbf{x}}$) is a solution in the core which contradicts the presumption we made and we show that proposition \textbf{(i)} holds.
                        
                        Then, we aim to show that \textbf{(ii)} \textit{In an ECG, with an efficient payoff distribution scheme, the objective is $\mathbf{\max_{\pi} v^{\pi}(\{\mathcal{N}\})}$.}
                        
                        The objective of a CG is finding a solution in the core. According to \textbf{(i)}, $\max_{\mathcal{CS} \in \mathcal{CS}_{\mathcal{N}}} v^{*}(\mathcal{CS})$ is equivalent to finding a solution in the core corresponding to the grand coalition $\{\mathcal{N}\}$. For this reason, we can write
                        \begin{align}
                            \max_{\mathcal{CS} \in \mathcal{CS}_{\mathcal{N}}} v^{*}(\mathcal{CS}) &= v^{*}(\{\mathcal{N}\}) \notag\\
                            &= \max_{\pi_{\mathcal{N}}} v^{\pi_{\mathcal{N}}}(\{\mathcal{N}\}) \notag\\
                            &\text{(Since we write $\pi = \pi_{\mathcal{N}}$)} \notag\\
                            &= \max_{\pi} v^{\pi}(\{\mathcal{N}\}). \label{eq:equiv}
                        \end{align}
                        Therefore, we prove \textbf{(ii)}.
                        
                        According to \textbf{(ii)}, we can conclude that in an ECG, the objective is maximizing $v^{\pi}(\{\mathcal{N}\})$, i.e., the global reward. However, an efficient payoff distribution scheme, e.g., Shapley value should be a condition, otherwise, $\max_{\pi} v^{\pi}(\{\mathcal{N}\}) < \hat{v}(\{\mathcal{N}\})$, where $\hat{v}(\{\mathcal{N}\})$ is the theoretically optimal value that can be found.
                    \end{proof}
                \end{theorem}
            \addtocounter{theorem}{-1}
        \endgroup
    
    \subsection*{Experimental Setups}
    \label{appendix:experiment settings}
        As for the setups of experiments, because different environments may involve variant complexity and dynamics, we give different hyperparameters for each task. All of algorithms use MLPs as hidden layers for the policy networks. All of policy networks only use one hidden layer. About the critic networks, every algorithm uses MLPs with one hidden layer. For each experiment, we keep the learning rate, entropy regularization coefficient, update frequency, batch size and the number of hidden units identical on each algorithm, except for the algorithms with the natural gradients (e.g., COMA and A2C). These algorithms need the special learning rates to maintain the stability of training. In the experiments, each agent has its own state in execution for the policy. In training, the agents with the centralised critics share the states while those with the decentralised critics only observe its own state. The rest details of experimental setups are introduced as below. All of models are trained by Adam Optimizer \cite{kingma2014adam} with default hyperparameters (except for the learning rate).
    
        \subsubsection{Additional Details of Cooperative Navigation}
        \label{appendix:cn_details}
            The specific hyperparameters of each algorithm used in Cooperative Navigation are shown as Tab.\ref{tab:hyperparameters_cn}. 
            
            \begin{table*}[ht!]
            \centering
            \scalebox{1.0}{
                \begin{tabular}{lcl}
                \toprule
                \textbf{Hyperparameters}               & \textbf{\#} & \textbf{Description}                                                \\ 
                \midrule
                hidden units                           & 32            & The number of hidden units for both policy and critic network \\
                training episodes                      & 5000          & The number of training episodes                               \\
                episode length                         & 200           & Maximum time steps per episode                                \\
                discount factor                        & 0.9           & The importance of future rewards                              \\
                update frequency for behaviour network & 100           & Behaviour network updates every \# steps                      \\
                learning rate for policy network       & 1e-4          & Policy network learning rate                                  \\
                learning rate for policy network(COMA)  & 1e-2          & Policy network learning rate for COMA                                 \\
                learning rate for policy network(IA2C)  & 1e-6          & Policy network learning rate for IA2C                                 \\                
                learning rate for critic network       & 1e-3          & Critic network learning rate                                  \\
                learning rate for critic network(COMA)  & 1e-4          & Critic network learning rate for COMA                                 \\
                learning rate for critic network(IA2C)  & 1e-5          & Critic network learning rate for IA2C                                 \\
                update frequency for target network    & 200           & Target network updates every \# steps                         \\
                target update rate                     & 0.1           & Target network update rate                                    \\
                entropy regularization coefficient     & 1e-2          & Weight or regularization for exploration                      \\
                batch size                             & 32            & The number of  transitions for each update                    \\ 
                \bottomrule
                \end{tabular}}
                \setlength{\abovecaptionskip}{10pt} 
                \caption{\small{Table of hyperparameters for Cooperative Navigation.}}
                \label{tab:hyperparameters_cn}
            \end{table*}
        
        \subsubsection{Additional Details of Prey-and-Predator} 
\begin{figure*}[t!]
                \centering
                \begin{subfigure}[b]{\textwidth}
                    \centering
                    \includegraphics*[scale=0.29]{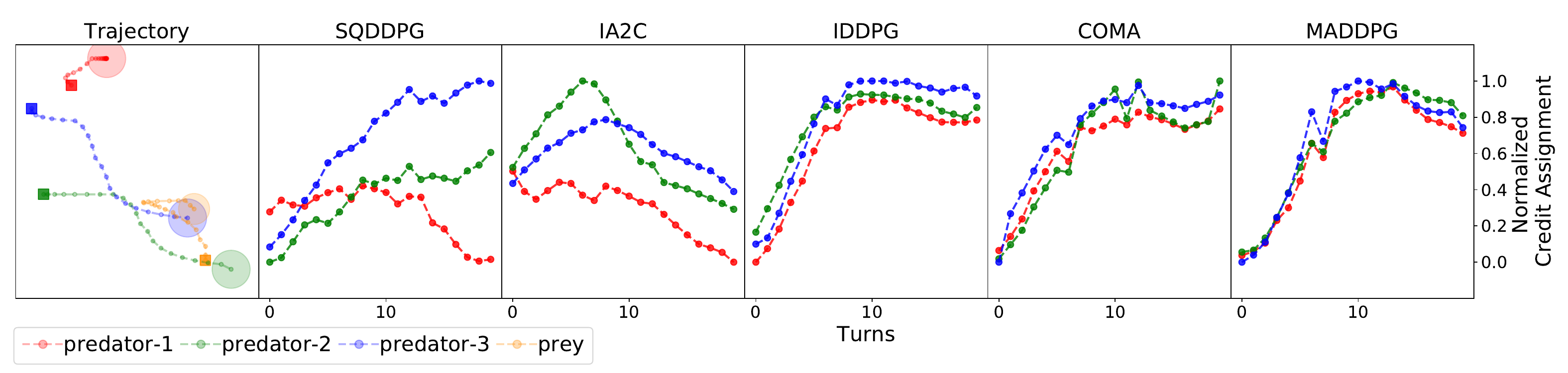}
                \end{subfigure}%
                \\
                \begin{subfigure}[b]{\textwidth}
                    \centering
                    \includegraphics*[scale=0.29]{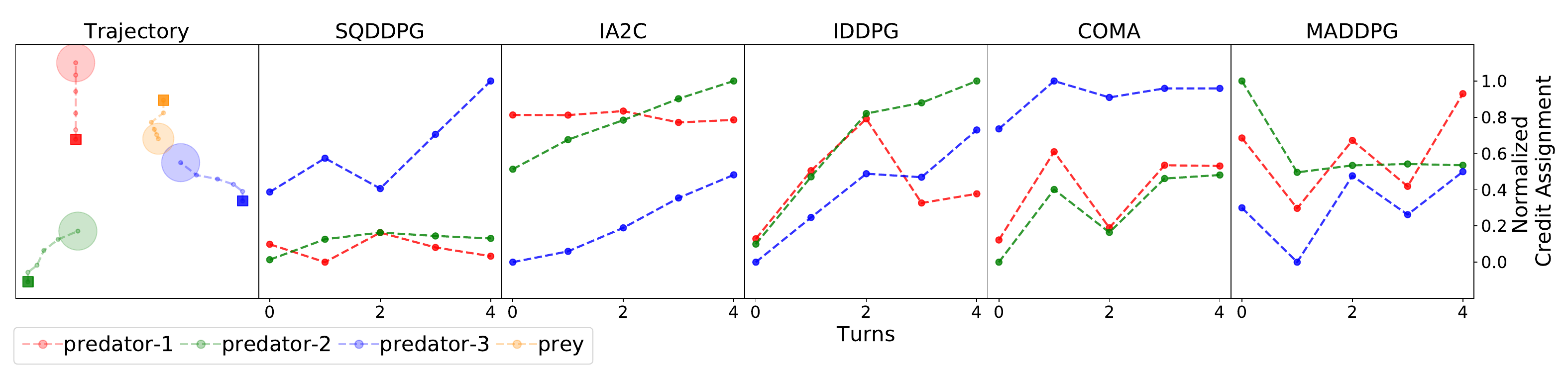}
                \end{subfigure}
                \\
                \begin{subfigure}[b]{\textwidth}
                    \centering
                    \includegraphics*[scale=0.29]{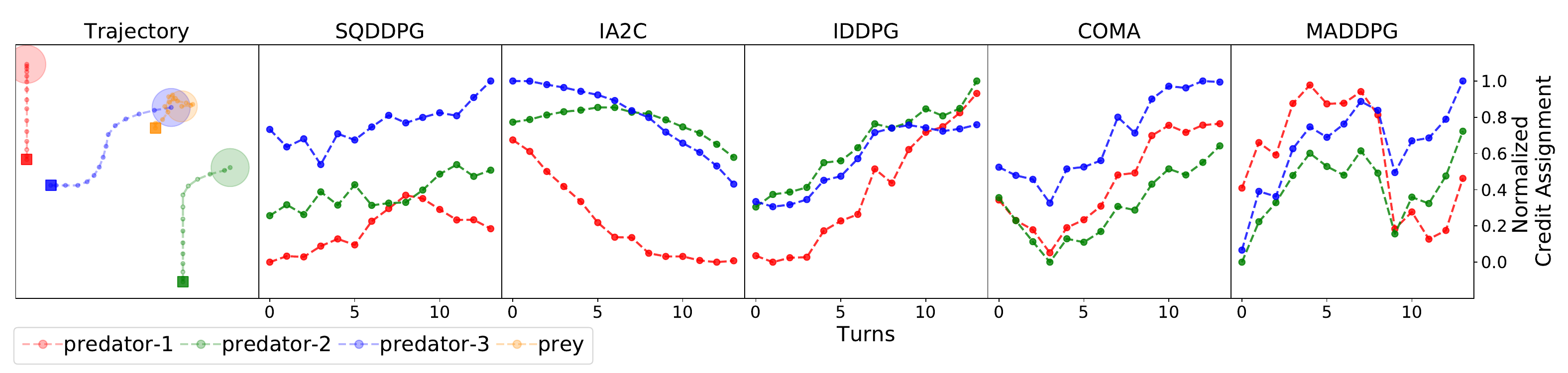}
                \end{subfigure}
                \\
                \begin{subfigure}[b]{\textwidth}
                    \centering
                    \includegraphics*[scale=0.29]{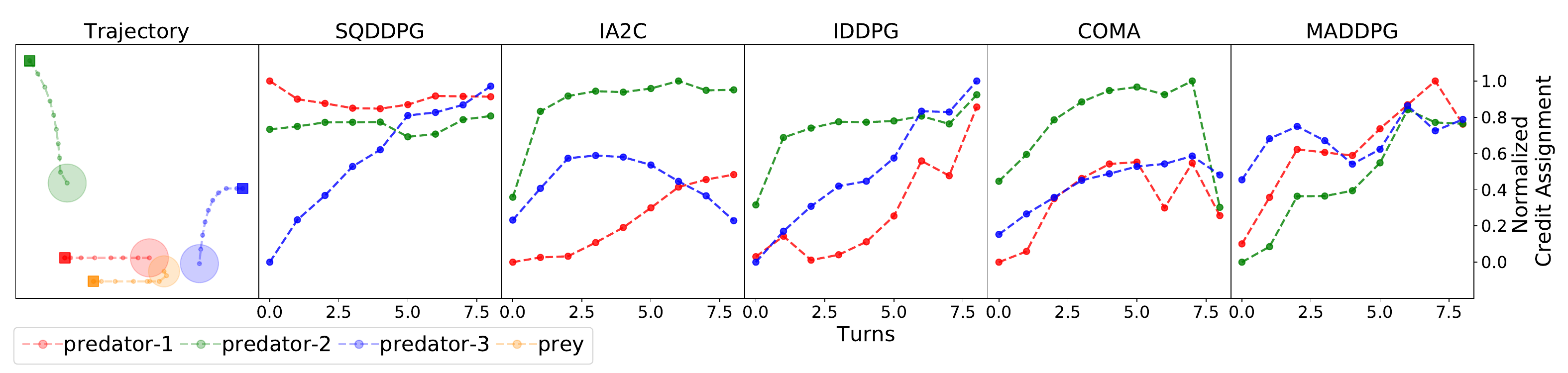}
                \end{subfigure}
                \caption{Credit assignment to each predator for a fixed trajectory. The leftmost figure records a trajectory sampled by an expert policy. The square represents the initial position whereas the circle indicates the final position of each agent. The dots on the trajectory indicates each agent's temporary positions. The other figures show the normalized credit assignments generated by different MARL algorithms according to this trajectory.}
                \label{fig:credit_extra}
            \end{figure*}

The specific hyperparameters of each algorithm used in Prey-and-Predator are shown as Tab.\ref{tab:hyperparameters_pp}. Also, we provide more evidences to support the conclusion that the credit assignment of SQDDPG is inversely proportional to the distance to the prey, shown as Fig.\ref{fig:credit_extra}.
            
            \begin{table*}[ht!]
            \centering
            \scalebox{1.0}{
                \begin{tabular}{lcl}
                \toprule
                \textbf{Hyperparameters}               & \textbf{\#} & \textbf{Description}                                                \\ 
                \midrule
                hidden units                           & 128           & The number of hidden units for both policy and critic network \\
                training episodes                      & 5000          & The number of training episodes                               \\
                episode length                         & 200           & Maximum time steps per episode                                \\
                discount factor                        & 0.99          & The importance of future rewards                              \\
                update frequency for behaviour network & 100           & Behaviour network updates every \# steps                      \\
                learning rate for policy network       & 1e-4          & Policy network learning rate                                  \\
                learning rate for policy network(COMA/IA2C)  & 1e-3    & Policy network learning rate for COMA and IA2C                                 \\
                learning rate for critic network       & 5e-4          & Critic network learning rate                                  \\
                learning rate for critic network(COMA/IA2C)  & 1e-4    & Critic network learning rate for COMA and IA2C                                 \\
                update frequency for target network    & 200           & Target network updates every \# steps                         \\
                target update rate                     & 0.1           & Target network update rate                                    \\
                entropy regularization coefficient     & 1e-3          & Weight or regularization for exploration                      \\
                batch size                             & 128           & The number of  transitions for each update                    \\ 
                \bottomrule
                \end{tabular}}
                \setlength{\abovecaptionskip}{10pt} 
                \caption{Table of hyperparameters for Prey-and-Predator.}
                \label{tab:hyperparameters_pp}
            \end{table*}
    
    \subsubsection*{Additional Details of Traffic Junction} 
\begin{figure*}[ht!]
            \centering
            \begin{subfigure}[b]{0.32\textwidth}
                \includegraphics*[width=\textwidth]{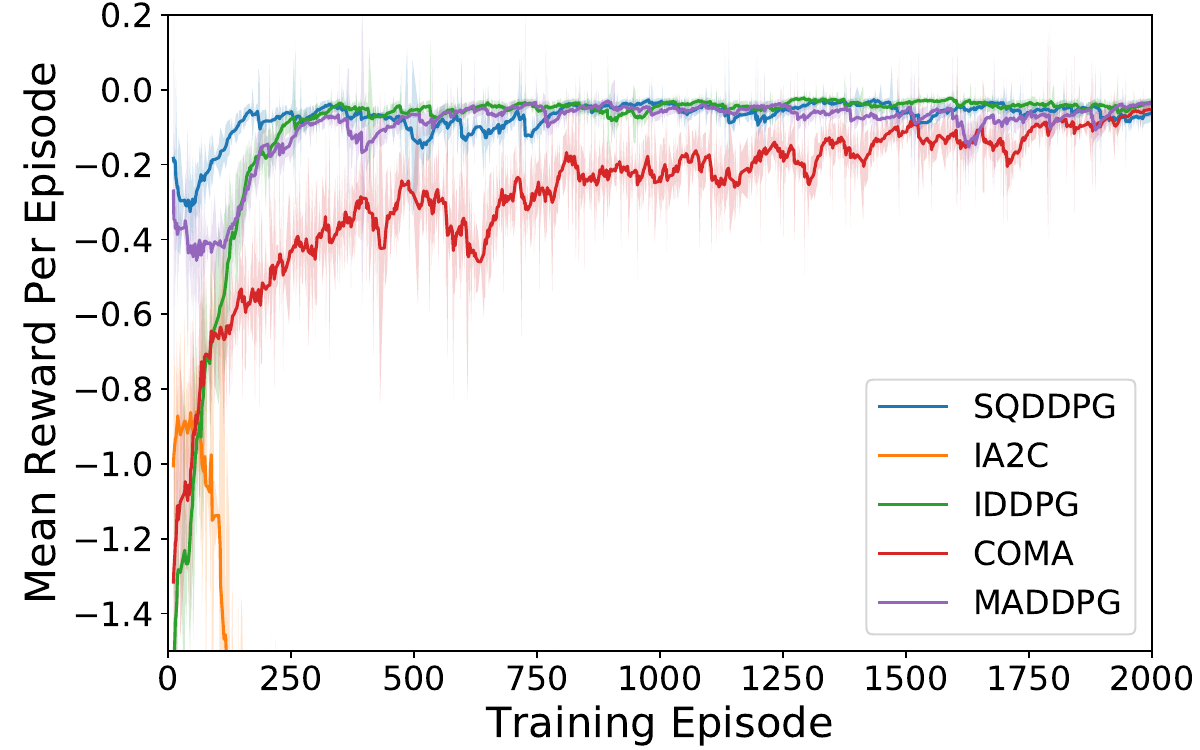}
                \caption{Mean reward per episode during training in Traffic Junction on easy version.}
                \label{fig:traffic_easy_reward}
            \end{subfigure}
            \begin{subfigure}[b]{0.32\textwidth}
                \includegraphics*[width=\textwidth]{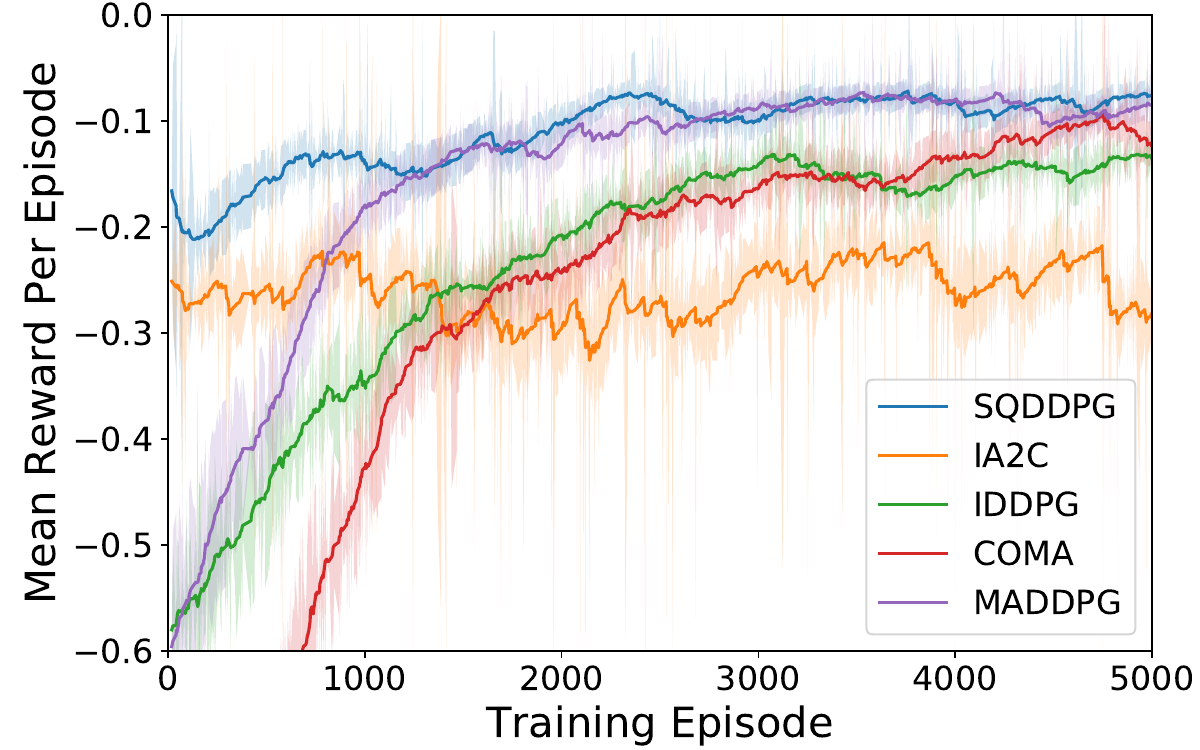}
                \caption{Mean reward per episode during training in Traffic Junction on medium version.}
                \label{fig:traffic_medium_reward}
            \end{subfigure}
            \begin{subfigure}[b]{0.32\textwidth}
                \includegraphics*[width=\textwidth]{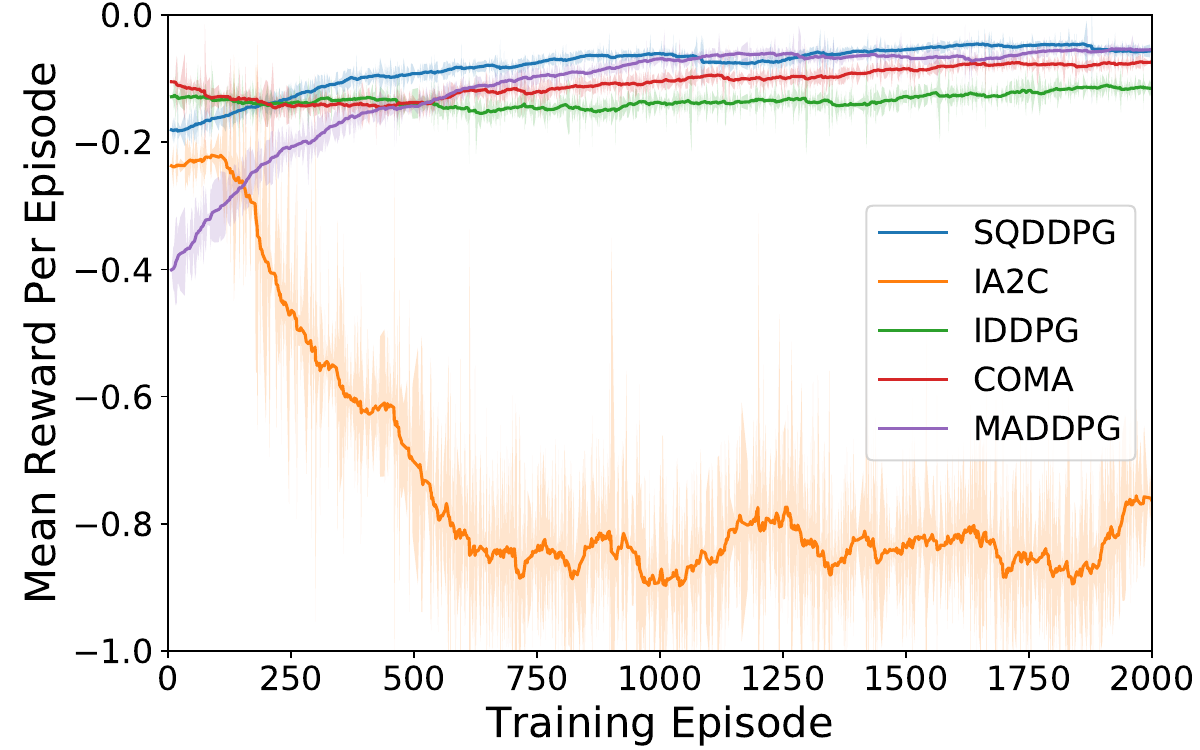}
                \caption{Mean reward per episode during training in Traffic Junction on high version.}
                \label{fig:traffic_hard_reward}
            \end{subfigure}
            \quad
            \begin{subfigure}[b]{0.32\textwidth}
                \includegraphics*[width=\textwidth]{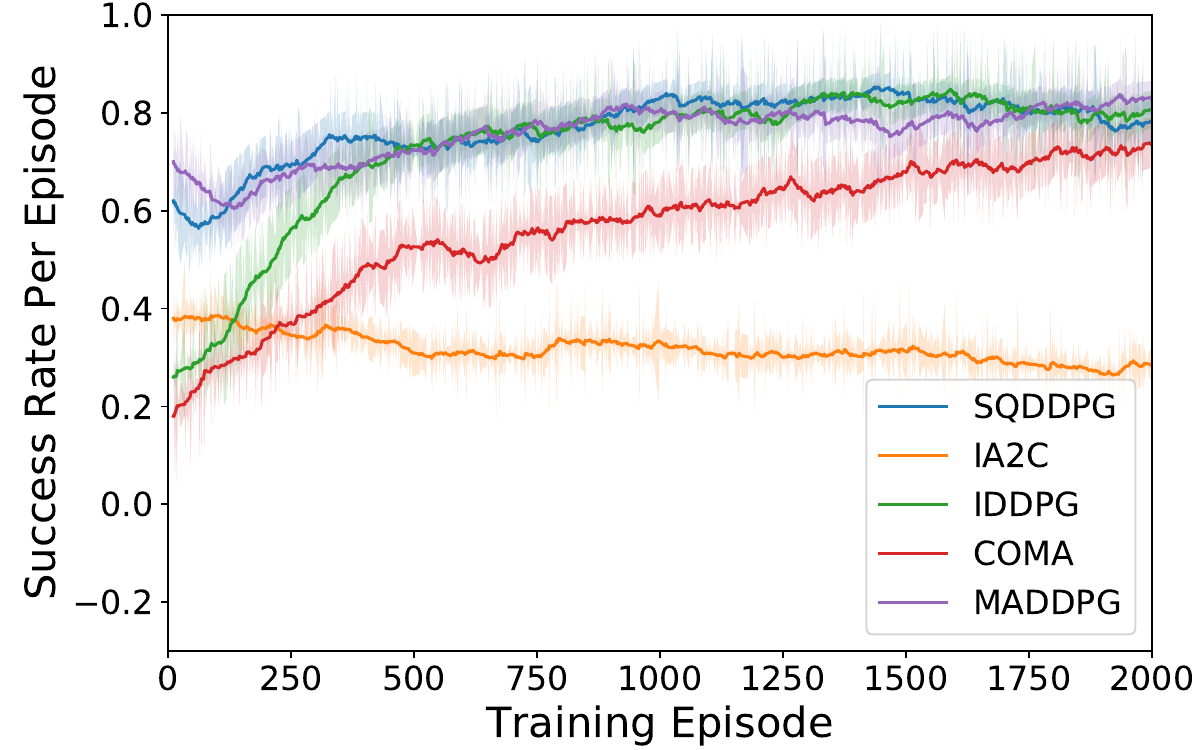}
                \caption{Success rate per episode during training in Traffic Junction on easy version.}
                \label{fig:traffic_easy_success}
            \end{subfigure}
            \begin{subfigure}[b]{0.32\textwidth}
                \includegraphics*[width=\textwidth]{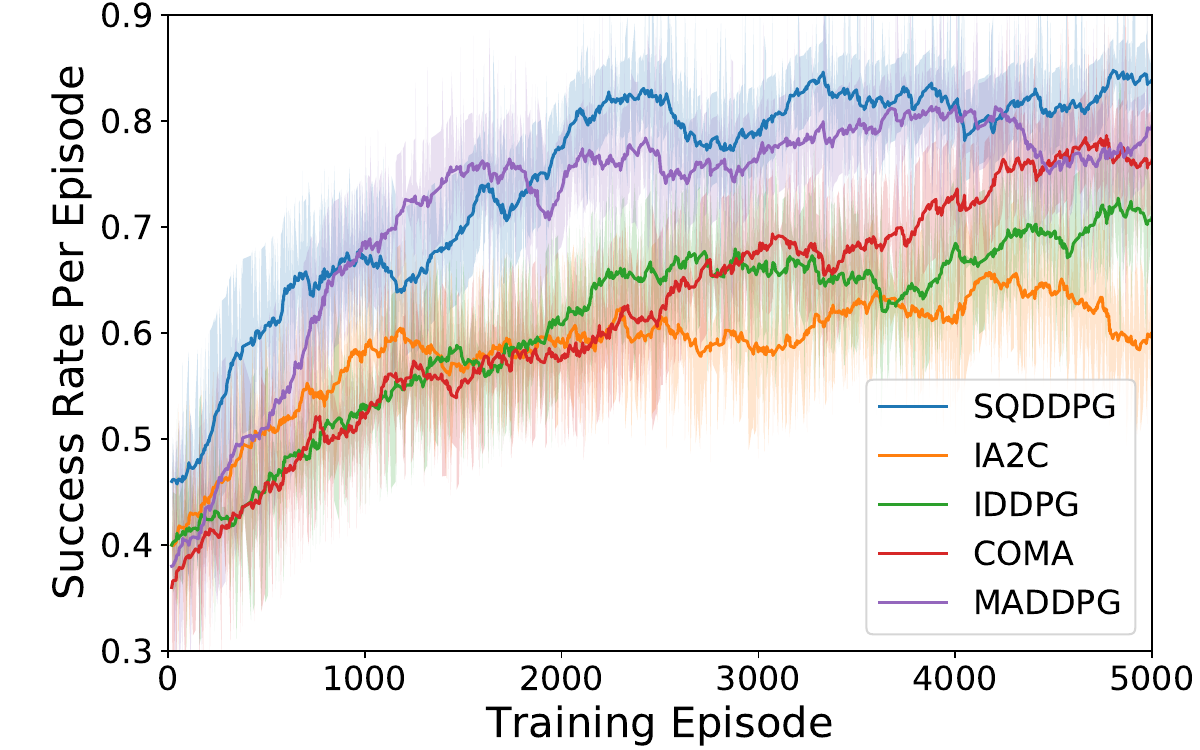}
                \caption{Success rate per episode during training in Traffic Junction on medium version.}
                \label{fig:traffic_medium_success}
            \end{subfigure}
            \begin{subfigure}[b]{0.32\textwidth}
                \includegraphics*[width=\textwidth]{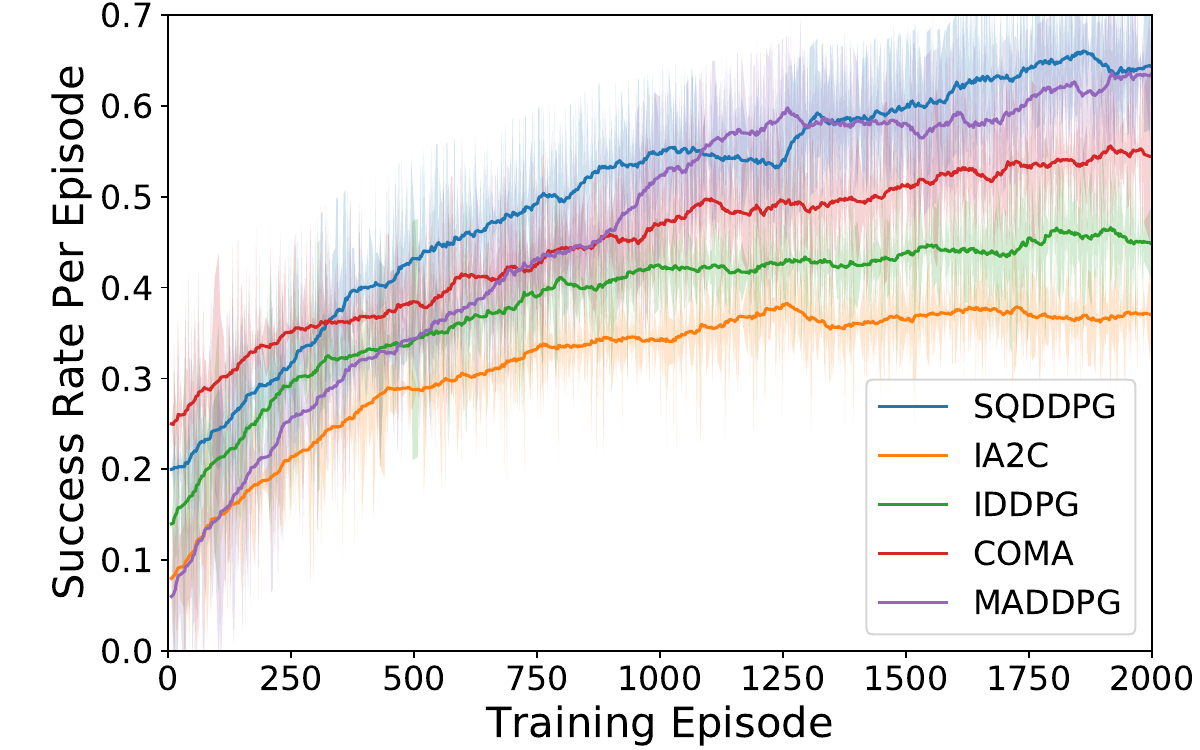}
                \caption{Success rate per episode during training in Traffic Junction on hard version.}
                \label{fig:traffic_hard_success}
            \end{subfigure}
            \caption{Mean reward and success rate per episode during training in Traffic Junction Environment on all difficulty levels.}
            \label{fig:traffic_junction_curve}
        \end{figure*}

\begin{figure*}[ht!]
            \centering
            \begin{subfigure}[b]{0.27\textwidth}
                \includegraphics[width=\textwidth]{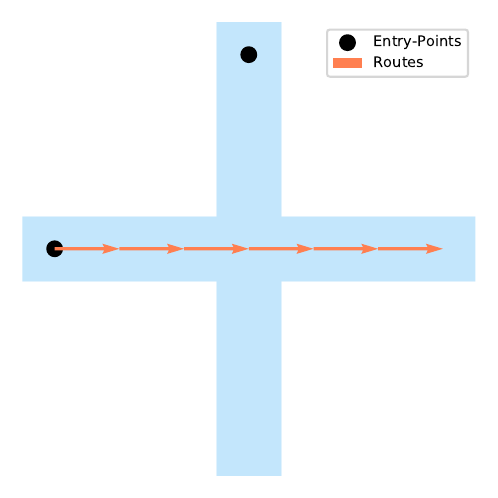}
                \caption{Easy}
                \label{fig:easy_env}
            \end{subfigure}
            \begin{subfigure}[b]{0.27\textwidth}
                \includegraphics[width=\textwidth]{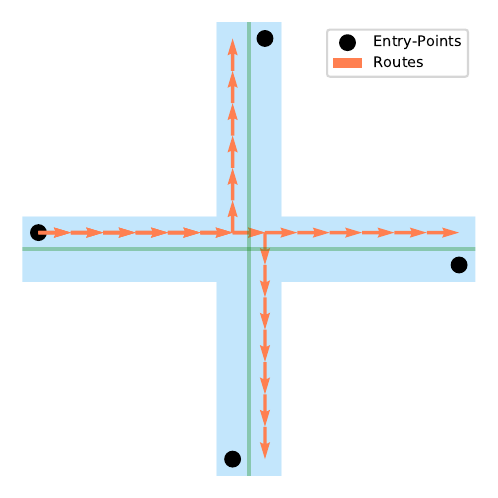} 
                \caption{Medium}
                \label{fig:medium_env}
            \end{subfigure}
            \begin{subfigure}[b]{0.27\textwidth}
                \includegraphics[width=\textwidth]{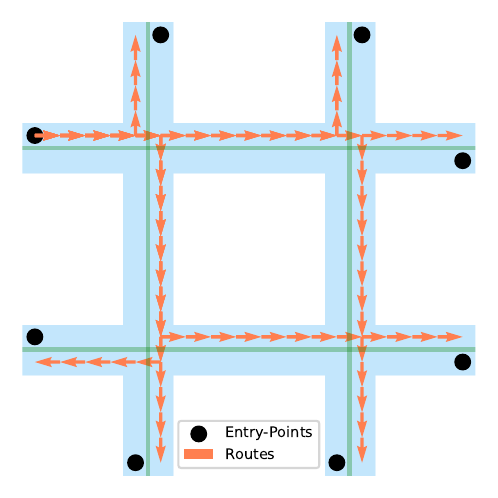}
                \caption{Hard}
                \label{fig:hard_env}
            \end{subfigure}
            \caption{Visualizations of traffic junction environment. The black points represent the available entry points. The orange arrows represent the available routes at each entry point. The green lines separate the two-way roads.}
            \label{fig:tf_roads}
        \end{figure*}

To give the reader an intuitive understanding of the environment, we list the experimental settings of different difficulty levels as Tab.\ref{tab:tj_setting} shows and give the illustrations shown as Fig.\ref{fig:tf_roads}. The specific hyperparameters of each algorithm used in Traffic Junction are shown as Tab.\ref{tab:hyperparameters_tj}. To exhibit the training procedure in more details, we also display the figures of mean rewards, e.g., Fig.\ref{fig:traffic_easy_reward}$\sim$\ref{fig:traffic_hard_reward} and the figures of success rate, e.g., Fig.\ref{fig:traffic_easy_success}$\sim$\ref{fig:traffic_hard_success}.
        
        \begin{table*}[ht!]
        \centering
        \scalebox{1.0}{
            \begin{tabular}{cccccccc}
                \toprule
                \textbf{Difficulty}  & \textbf{$p_{\text{arrive}}$} & \textbf{$N_{\max}$} & \textbf{Entry-Points \#} & \textbf{Routes \#} &  \textbf{Two-way} &\textbf{Junctions \#} & \textbf{Dimension} \\ 
                \midrule
                Easy  & 0.3  &  5  & 2  & 1 & F & 1 & 7x7 \\
                Medium & 0.2 &  10  & 4  & 3 & T & 1 & 14x14 \\
                Hard & 0.05  &  20  & 8  & 7 & T & 4 & 18x18 \\
                \bottomrule
            \end{tabular}}
            \caption{The settings of Traffic Junction for different difficulty levels. \textbf{$p_{\text{arrive}}$} means the probability to add an available car into the environment. \textbf{$N_{\max}$} means the existing number of the cars. \textbf{Entry-Points \#} means the number of possible entry points for each car. \textbf{Routes \#} means the number of possible routes starting from every entry point.}
            \label{tab:tj_setting}
        \end{table*}

        \begin{table*}[ht!]
        \centering
        \scalebox{0.92}{
            \begin{tabular}{lcccl}
            \toprule
            \textbf{Hyperparameters}               & \textbf{Easy} & \textbf{Meidum} & \textbf{Hard} & \textbf{Description}                                                \\ 
            \midrule
            hidden units                           & 128           & 128             & 128           & The number of hidden units for both policy and critic network \\
            training episodes                      & 2000          & 5000            & 2000          & The number of training episodes                               \\
            episode length                         & 50            & 50              & 100           & Maximum time steps per episode                                \\
            discount factor                        & 0.99          & 0.99            & 0.99          & The importance of future rewards                              \\
            update frequency for behaviour network & 25            & 25              & 25            & Behaviour network updates every \# steps                      \\
            learning rate for policy network       & 1e-4          & 1e-4            & 1e-4          & Policy network learning rate                                  \\
            learning rate for critic network       & 1e-3          & 1e-3            & 1e-3          & Critic network learning rate                                  \\
            update frequency for target network    & 50            & 50              & 50            & Target network updates every \# steps                         \\
            target update rate                     & 0.1           & 0.1             & 0.1           & Target network update rate                                    \\
            entropy regularization coefficient     & 1e-4          & 1e-4            & 1e-4          & Weight or regularization for exploration                      \\
            batch size                             & 64            & 32              & 32            & The number of  transitions for each update                    \\ 
            \bottomrule
            \end{tabular}}
            \setlength{\abovecaptionskip}{10pt} 
            \caption{Table of hyperparameters for Traffic Junction.}
            \label{tab:hyperparameters_tj}
        \end{table*}
 
    \subsection*{Non-cooperative Game Theory}
    \label{appdix:non-cooperative game theory}
        To clarify why we are interested in the framework of cooperative game theory, let us begin with non-cooperative game theory. Non-cooperative game theory aims to solve out the problem in which agents are selfish and intelligent \cite{osborne1994course}. In other words, each agent merely considers maximizing his own rewards to approach a Nash equilibrium. Consequently, non-cooperative game theory is intuitively not suitable for modelling cooperative scenarios. To model the cooperative scenarios, one possible way is through constructing a global reward function (which is unnecessary to be convex), e.g., Potential \cite{monderer1996potential} that replaces each individual reward function. As a result, each agent's objective is forced to be identical, i.e., fully cooperative game. Even if each agent is still keen on maximizing his own rewards, the coordination can be formed. Nonetheless, this approach has its own limitations. Firstly, it lacks the clear explanations for credit assignments. Secondly, assigning credits to each agent equally may cause the slow learning rate \cite{balch1997learning,balch1999reward}. Thirdly, this framework is difficult to be extended to solve out more complex problems such as the competition among different coalitions. However, cooperative game theory (with transferable utility), concentrating on credit assignment and dividing coalitions (or groups) \cite{chalkiadakis2011computational} can solve out these problems. This is the reason why we introduce and investigate cooperative game theory.
        
    \subsection*{Limitations of Extended Convex Game}
    \label{appdix:limitaion}
        In this paper, we propose a framework built on cooperative game theory called extended convex game (ECG). Although ECG has extended the framework of global reward game defined upon non-cooperative game theory to a broader scope, there exist some limitations to this model. Firstly, we have to assume that there is an oracle scheduling the coalition initially, however, this oracle is difficult to realize in implementation. Even if the oracle can be implemented, this model still cannot solve out some problems with random perturbations. This is due to the fact that the oracle has assigned each agent to a coalition with the environment that it knows. Obviously, the perturbation exceeds its knowledge. To deal with this problem, we may investigate how to enable the coalition construction dynamically in the future work. The intuitive idea is enabling the oracle to learn a policy for scheduling the coalition from the history information. At each step, it uses the learned policy to divide the coalitions. Then, each agent act within the coalition to maximize the social value of the coalition. This process can be repeated infinitely. Nonetheless, the promising convergence under the cooperative game theoretical framework for this complicated process could be a challenge.
        
%===============================================================================

\bibliographystyle{aaai}
{\small\bibliography{aaai}}

%===============================================================================
        
\end{document}